\documentclass[journal]{IEEEtran}
\usepackage[para]{threeparttable}
\usepackage{amssymb,amsmath}
\usepackage[pagebackref=false,colorlinks]{hyperref}
\usepackage{cite}
\usepackage{hyperref}   
\usepackage{amsmath}  
\usepackage{booktabs}
\usepackage{multirow}
\usepackage{caption}%
\usepackage{microtype}%
\usepackage{verbatim}%
\usepackage{bm} %
\usepackage{indentfirst}%
\usepackage{graphicx}
\usepackage{float}
\usepackage{rotate}
\usepackage[section]{placeins}  
\usepackage{setspace}  
\usepackage{paralist}
\usepackage{colortbl}
\usepackage{utfsym}
\usepackage{color}
\usepackage{subfigure}
\usepackage{makecell}

\title{FA-Depth: Toward Fast and Accurate Self-supervised Monocular Depth Estimation}
\author{Fei~Wang,~\IEEEmembership{Student Member,~IEEE}, Jun~Cheng,~\IEEEmembership{Senior Member,~IEEE}
\vspace{-15pt}
\thanks{\newline\indent Corresponding author: Jun Cheng \newline\indent Fei~Wang, Jun~Cheng are with the Guangdong-Hong Kong-Macao Joint Laboratory of Human-Machine Intelligence-Synergy Systems, Shenzhen Institute of Advanced Technology, Chinese Academy of Sciences, Shenzhen, 518055, China. Fei~Wang is also with the Shenzhen College of Advanced Technology, University of Chinese Academy of Sciences, Beijing 100049, China, and also with the Department of Mechanical and Automation Engineering, The Chinese University of Hong Kong, Hong Kong. (email:\{fei.wang2, jun.cheng\}@siat.ac.cn).}}

\begin{document}
\maketitle
\begin{abstract}
Most existing methods often rely on complex models to predict scene depth with high accuracy, resulting in slow inference that is not conducive to deployment. To better balance precision and speed, we first designed SmallDepth based on sparsity. Second, to enhance the feature representation ability of SmallDepth during training under the condition of equal complexity during inference, we propose an equivalent transformation module(ETM). Third, to improve the ability of each layer in the case of a fixed SmallDepth to perceive different context information and improve the robustness of SmallDepth to the left-right direction and illumination changes, we propose pyramid loss. Fourth, to further improve the accuracy of SmallDepth, we utilized the proposed function approximation loss (APX) to transfer knowledge in the pretrained HQDecv2, obtained by optimizing the previous HQDec to address grid artifacts in some regions, to SmallDepth. Extensive experiments demonstrate that each proposed  component improves the precision of SmallDepth without changing the complexity of SmallDepth during inference, and the developed approach achieves state-of-the-art results on KITTI at an inference speed of more than 500 frames per second and with approximately 2 M parameters. The code and models will be publicly available at \href{https://github.com/fwucas/FA-Depth}{FA-Depth}.

\end{abstract}

\IEEEpeerreviewmaketitle

\section{Introduction}\label{sec:intro}

\IEEEPARstart {S}{cene} depth not only plays an important role in robotics \cite{yasuda2020autonomous}, autonomous driving \cite{chuah2021deep} and augmented reality but can also be used as auxiliary information to assist in improving other tasks (e.g., object detection \cite{chen2020improved,zhao2022joint}). Although the existing methods \cite{li2021revisiting,chuah2021deep,eigen2014depth,liu2015learning,yuan2022neural,fu2018deep,fu2018deep,ranftl2021vision,bhat2021adabins,patil2022p3depth,shin2023deep,li2023learning,yang2023gedepth,yin2021virtual,wang2023joint,wei2024fs} have achieved competitive performance in dense depth estimation, these methods depend on the availability of large-scale dense per-pixel depth annotations or even require stereo video \cite{li2021revisiting} sequences for network training. However, massive amounts of high-quality labeled data are expensive and impractical to acquire. Alternatively, self-supervised monocular depth estimation methods \cite{godard2019digging,bian2019unsupervised,wang2022cbwloss,bello2021self,wang2023planedepth,bangunharcana2023dualrefine,he2022ra,garg2016unsupervised,zhou2017unsupervised,lyu2021hr,guizilini20203d,song2021mlda,zhang2020unsupervised,wangfei_hqdec,wang2022unsupervised,miao2023ds,feng2023iterdepth,liu2023self,wu2023self,zhao2023sptr} are increasingly favored by researchers. To improve the accuracy of self-supervised monocular depth estimation, more prior knowledge \cite{godard2019digging,bian2019unsupervised,wang2022cbwloss,bello2021self,wang2023planedepth,bangunharcana2023dualrefine,he2022ra}, additional tasks \cite{zhang2021dpsnet,ranjan2019competitive,sun2021unsupervised,wang2020adversarial}, multiframe input \cite{guizilini2022multi,watson2021temporal,bangunharcana2023dualrefine,wang2023crafting,miao2023ds,feng2023iterdepth,wu2023self}, or  semantic information \cite{klingner2020self,jung2021fine,xu2021multi} were employed to provide more accurate directions for updating model parameters. More advanced network architectures \cite{lyu2021hr,guizilini20203d,song2021mlda,zhang2020unsupervised,wangfei_hqdec,wang2022unsupervised,zhao2023sptr} were also designed to improve the feature representation capability of DepthNet.

However, the existing methods still have many shortcomings. First, DepthNet, which has high complexity in most previous works \cite{wangfei_hqdec,zhang2020unsupervised,song2021mlda,guizilini20203d,lyu2021hr,he2022ra,wang2022cbwloss,bian2019unsupervised,godard2019digging,zhao2023sptr}, was usually utilized to achieve higher accuracy, resulting in slow speed during inference. Although there have also been a few works \cite{poggi2022real,peluso2021monocular,zhang2023lite,song2023spatial}  in which lightweight DepthNet was designed to improve the inference speed, this comes with the cost of greatly reducing the number of feature maps or stages. To this end, we designed SmallDepth based on sparsity to better balance precision and speed. 

Second, during training and inference, the same network architectures for DepthNet as the existing works \cite{wangfei_hqdec,zhang2020unsupervised,song2021mlda,guizilini20203d,lyu2021hr,he2022ra,wang2022cbwloss,bian2019unsupervised,godard2019digging,poggi2022real,peluso2021monocular,zhang2023lite,song2023spatial,zhao2023sptr} were employed, resulting in the feature representation ability of the model being limited during training under the condition of the same complexity during inference. To this end, we propose the ETM. 

Third, during each iteration update of the training period, the above methods \cite{wangfei_hqdec,zhang2020unsupervised,song2021mlda,guizilini20203d,lyu2021hr,he2022ra,wang2022cbwloss,bian2019unsupervised,godard2019digging,poggi2022real,peluso2021monocular,zhang2023lite,song2023spatial,zhao2023sptr} can only use the information contained in the source image to calculate the supervisory signal to guide the update of the model weight. To this end, we propose the pyramid loss. 

Fourth, although previous work \cite{wangfei_hqdec} has achieved excellent performance, grid artifacts exist in some regions of the predicted depth map. To this end, we improved HQDec \cite{wangfei_hqdec}, resulting in a new version, named HQDecv2. 

Finally, (a) only pseudo-depth in previous works \cite{petrovai2022exploiting,sun2023sc,shi20233d,li2023sense,lyu2021hr} was employed as a supervisory signal for the lightweight model. (b) In the above works, either the pseudo-depth generated by the pretrained large model based on self-supervised learning was assumed to be correct by default \cite{lyu2021hr}, only the false prediction depth map caused by the occluded area was filtered out \cite{petrovai2022exploiting}, the false prediction caused by the reflective surfaces was processed by using the ground pose \cite{shi20233d}, the pretraining network based on ground truth was used to generate pseudo-depth \cite{sun2023sc}. To this end, we propose APX.

Specifically, we redesigned the three main components (downsampling, feature encoding, and upsampling) of DepthNet based on the analysis of the factors influencing of the model complexity index according to formula \eqref{eq_complexity} and obtained SmallDepth. Second, in the proposed ETM, we utilized randomly dropped filters with different shapes in parallel to model different context information during training, while only a single-shape filter was utilized to propagate the information flow during inference. Third, we utilized the proposed pyramid loss to (a) enforce each layer of fixed model architecture to perceive different context information with consistency and (b) improve the robustness of the model to left-right and illumination changes by simultaneously measuring the photometric difference of both source and the corresponding transformed (e.g., scaled, flipped, color enhanced) images and enforcing the consistency between the corresponding feature/depth maps between them. Fourth, we replaced the original modules  \cite{wangfei_hqdec} with the corresponding improved version (see Sec. \ref{sec:hqdecv2_method}) to address grid artifacts and reduce complexity. Finally, to further improve the accuracy of SmallDepth, we utilized the masked probability distribution difference between the outputs (e.g., feature/disparity maps obtained by small and large models, respectively) as an additional supervisory signal to guide the updating of the small model(e.g., SmallDepth), making its distribution characteristics similar to those of the large model(e.g., HQDecv2).

Our main contributions are summarized as follows:(1) To better balance precision and speed, we designed SmallDepth based on sparsity. (2) To increase the feature representation ability of the model without changing the complexity of the model during inference, we proposed the ETM. (3) To improve the ability of each layer in the case of a fixed SmallDepth to perceive different context information and improve the robustness of the SmallDepth to the left-right direction and illumination changes, we proposed pyramid loss. (4) To further improve the accuracy of SmallDepth, we utilized the proposed APX to transfer knowledge in the pretrained HQDecv2, obtained by optimizing the previous HQDec to deal with grid artifacts in some regions, to SmallDepth.

\vspace{-10pt}
\section{Related work}\label{sec:relatedwork}

\subsection{Supervised Depth Estimation}

The supervised depth estimation algorithm relies on manually labeled sparse 3D point clouds to train DepthNet. In 2014, Eigen et al. \cite{eigen2014depth} first predicted a depth map from a single image by stacking coarse-scale and fine-scale networks. Subsequently, depth estimation was formulated as a deep continuous conditional random fields (CRFs) learning problem, but geometric priors were ignored \cite{liu2015learning}. Instead of plain CRFs, neural window fully-connected CRFs \cite{yuan2022neural} were proposed to reduce the computational complexity. To eliminate the influence of large depth values on the training process, the depth estimation task was cast as an ordinal regression problem by discretizing continuous depth into many intervals in log space \cite{fu2018deep} or utilizing ordinal regression with context information to capture fine depth features \cite{meng2021cornet}. To improve the ability of generalizing to other scenarios, Zhu et al. \cite{zhu2023ha} utilized hierarchical adaptive bins to learn diverse scene priors from multiple datasets. To model a longer range of context information, attention mechanisms \cite{vaswani2017attention}, in particular, the vision transformer (ViT) \cite{dosovitskiy2020image}, have also been used to achieve improved depth estimation \cite{ranftl2021vision,bhat2021adabins}. For example, Ranftl et al. \cite{ranftl2021vision} leveraged ViT in place of convolutional neural networks (CNNs) as the encoder to achieve finer-grained and more globally coherent predictions, yielding substantial improvements. Bhat et al. \cite{bhat2021adabins} computed adaptive bins by replacing the scheme that divides the depth range into a fixed number of intervals \cite{fu2018deep} by using a transformer-based architecture block, resulting in a global depth map.

In addition, various prior knowledge (e.g., planarity priors \cite{patil2022p3depth,chuah2021deep}, thermal images \cite{shin2023deep}, multicue fusion \cite{li2023learning}, ground priors \cite{yang2023gedepth}, and virtual normal directions \cite{yin2021virtual}) or multitask learning \cite{wang2023joint} has also been utilized to improve the predicted depth.

\vspace{-10pt}
\subsection {Self-supervised Depth Estimation}
Although supervised models achieve excellent performance, they are not universally applicable and rely heavily on obtained ground truth data. Moreover, the data annotation process is often slow and expensive. The obtained annotations are also affected by structural artifacts, especially in the presence of reflective, transparent and dark surfaces or nonreflective sensors that output infinite values. All of these challenges strongly motivate us to infer depth in an unsupervised way. Based on the photometric difference, the depth map  was first successfully recovered from a single view in an unsupervised way \cite{garg2016unsupervised}, but the camera poses needed to be known. To release this limitation, Zhou et al. \cite{zhou2017unsupervised} first proposed a completely unsupervised monocular estimation method. However, the estimated depth maps are far inferior to those based on supervised methods.

To further improve the quality of depth maps based on unsupervised schemes, more prior knowledge \cite{godard2019digging,bian2019unsupervised,wang2022cbwloss,bello2021self,wang2023planedepth,bangunharcana2023dualrefine,he2022ra,miao2023ds} was mined. e.g., Bian et al. \cite{bian2019unsupervised} adopted a geometric consistency constraint to explicitly enforce scale consistency between different samples. Godard et al. \cite{godard2019digging} utilized an auto-mask to let the network ignore objects that move at the same velocity as the camera. Wang et al. \cite{wang2022cbwloss} utilized bidirectional camera flow to handle dynamic objects and occlusions and adopted feature perception loss to enhance the ability of DepthNet to perceive textureless regions without increasing overhead. Miao et al. \cite{miao2023ds} improved incorrectly occluded regions  in static cost volumes by exploiting residual optical flow to describe moving objects. Bello et al. \cite{bello2021self} further refined the estimated depth maps via confidence-guided data augmentation. Furthermore, vertical and ground planes \cite{wang2023planedepth} were also considered to obtain a much smoother depth for ground regions and assisted in identifying the drivable regions. In contrast to \cite{bello2021self}, Bangunharcana et al. \cite{bangunharcana2023dualrefine} utilized epipolar geometry constraints to iteratively refine the estimated depth map.  He et al. \cite{he2022ra} improved the robustness to different resolutions by explicitly learning the scale invariance between the source and enhanced image.

Similar to supervised methods, more advanced DepthNets \cite{lyu2021hr,guizilini20203d,song2021mlda,zhang2020unsupervised,wangfei_hqdec,wang2022unsupervised} have also been designed. e.g., different dense connection schemes \cite{zhang2020unsupervised,lyu2021hr,wangfei_hqdec,song2021mlda} between low-level and high-level features were designed to obtain depth maps with sharp details. The attention mechanism/ViT was introduced to improve the long-range modeling ability of DepthNet \cite{wangfei_hqdec,wang2022unsupervised}. To reduce the information loss caused by the downsampling process and recover important spatial information, packing-unpacking blocks \cite{guizilini20203d} and attention-based sampling modules \cite{wangfei_hqdec} were developed. In addition, dense structure information was preserved by empowering the transformer to effectively leverage long-range cross-modal depth dependencies \cite{zhao2023sptr}.

Different from \cite{godard2019digging,bian2019unsupervised,wang2022cbwloss}, to explicitly address dynamic objects and occlusions, additional subnetworks (e.g., segmentation networks \cite{zhang2021dpsnet,ranjan2019competitive}, masknets \cite{sun2021unsupervised}) were introduced. Instead of utilizing these methods to deal with dynamic scenes, Wang et al. \cite{wang2020adversarial} utilized one (namely, a discriminator) to force the consistency of latent representations between real and reconstructed views, yielding accurate depth and  ego-motion estimation. In addition, multi-frame \cite{guizilini2022multi,watson2021temporal,bangunharcana2023dualrefine,wang2023crafting,miao2023ds,feng2023iterdepth,wu2023self} or semantic information \cite{klingner2020self,jung2021fine,xu2021multi} was also utilized to obtain improved depth maps.
\vspace{-5pt}
\subsection {Lightweight Model for Depth Estimation}

Although the performance of depth estimation based on deep learning has been greatly improved in recent years, most methods have slow inference and high complexity, making them difficult to deploy in real-world scenarios. 

To address these challenges, lightweight models \cite{zhang2023lite,poggi2022real,song2023spatial,papa2023meter,peluso2021monocular} have attracted  researchers' attention. To better balance available resources and execution time, Poggi et al. \cite{poggi2022real}  designed a small pyramidal encoder with fewer feature maps at the expense of multiple decoders, aiming at reducing the memory footprint and runtime without reducing the overall accuracy. Based on \cite{poggi2022real}, Peluso et al. \cite{peluso2021monocular} maximized the saved resources by scaling the resolution and using a shallower pyramidal architecture to meet with microcontrollers. Although its speed improved, its accuracy was too low. Subsequently, an attention mechanism  (e.g., cross-covariance attention \cite{ali2021xcit}) was reasonably combined with CNNs to better balance speed and accuracy\cite{zhang2023lite,papa2023meter}. e.g., convolutions with smaller kernels were directly combined with fewer transformer blocks to reduce the computational complexity.

\vspace{-10pt}
\subsection{Knowledge Distillation for Depth Estimation}
To further improve the performance of the small model without increasing complexity, schemes based on knowledge distillation \cite{petrovai2022exploiting,sun2023sc,shi20233d,li2023sense,lyu2021hr} were utilized to transfer knowledge from a large model with high precision but slow inference to a small model with low precision but fast inference, thereby improving accuracy. Lyu et al. \cite{lyu2021hr} directly utilized pseudo-depth obtained by high-performance networks to train lightweight models.
Based on the pseudo-depth data generated by supervised training, Sun et al. \cite{sun2023sc} modeled the nearer/further relation of the estimated depth between dynamic and static regions by ranking relations between any two pixels to cope with the dynamic regions. Petrovai et al. \cite{petrovai2022exploiting} directly utilized the absolute pseudo-depth, obtained from a pretrained model and real camera height, to train a more lightweight network. To improve the depth prediction accuracy for reflective surfaces without increasing the computational cost, Shi et al. \cite{shi20233d} utilized the pseudo labels, obtained by fusing the predicted depth map from the pretrained model and the projected depth map based on the ground truth pose, as a supervisory signal to train the network. Different from \cite{petrovai2022exploiting,sun2023sc}, Li et al. \cite{li2023sense}  
utilized self-evolving pseudo labels, which can be iteratively updated in the form of a loop iteration, to progressively improve the performance of self-supervised monocular depth estimation. Liu et al. \cite{liu2023self} used a parameter-optimized model as the teacher updated as the training epochs to provide additional supervision during the training process, however, the teacher model had the same structure as the student model.

However, the above methods require ground truth information \cite{li2021revisiting,chuah2021deep,eigen2014depth,liu2015learning,yuan2022neural,fu2018deep,fu2018deep,ranftl2021vision,bhat2021adabins,patil2022p3depth,shin2023deep,li2023learning,yang2023gedepth,yin2021virtual,wang2023joint}, auxiliary networks \cite{zhang2021dpsnet,ranjan2019competitive,sun2021unsupervised}, semantic information \cite{klingner2020self,jung2021fine,xu2021multi} for network training, or even multiframe inputs \cite{guizilini2022multi,watson2021temporal,bangunharcana2023dualrefine,wang2023crafting} during inference. More importantly, the above methods either have high complexity \cite{wangfei_hqdec,zhang2020unsupervised,song2021mlda,guizilini20203d,lyu2021hr,he2022ra,wang2022cbwloss,bian2019unsupervised,godard2019digging}, resulting in slow inference speed, or increase inference speed by greatly reducing the number of feature maps or stages  \cite{poggi2022real,peluso2021monocular,zhang2023lite,song2023spatial}, resulting in low precision of depth estimation. Second, the same network architectures were used in the above methods during both training and inference \cite{wangfei_hqdec,zhang2020unsupervised,song2021mlda,guizilini20203d,lyu2021hr,he2022ra,wang2022cbwloss,bian2019unsupervised,godard2019digging,poggi2022real,peluso2021monocular,zhang2023lite,song2023spatial}, resulting in the feature representation ability of the model being limited during training under the condition of the same complexity during inference. Third, during each iterative update of the training period, the above methods \cite{wangfei_hqdec,zhang2020unsupervised,song2021mlda,guizilini20203d,lyu2021hr,wang2022cbwloss,bian2019unsupervised,godard2019digging,poggi2022real,peluso2021monocular,zhang2023lite,song2023spatial} only employ the supervisory signal calculated from source samples to guide the update of the model weight. Fourth, although previous work \cite{wangfei_hqdec} has achieved excellent performance, grid artifacts exist in some regions of the predicted depth map. Ultimately, (a) only  the pseudo-depth in previous works \cite{petrovai2022exploiting,sun2023sc,shi20233d,li2023sense,lyu2021hr} was employed as a supervisory signal for the lightweight model. (b) The pseudo-depth was assumed to be correct by default \cite{lyu2021hr}, only the false prediction depth map caused by the occluded area was filtered out \cite{petrovai2022exploiting}, the false prediction caused by the reflective surfaces was processed by using the ground pose \cite{shi20233d}, or the pretraining network based on the ground truth was used to generate the pseudo-depth \cite{sun2023sc}.

To cope with these challenges, in this paper, we (a) designed SmallDepth based on sparsity, (b) proposed ETM, (c) proposed pyramid loss, (d) improved HQDec, and (e) proposed APX.
\vspace{-5pt}
\section{Method}\label{sec:method}

\begin{figure*}[htbp]
	\centering 
	\vspace{-5pt}	
	\includegraphics[scale=0.52]{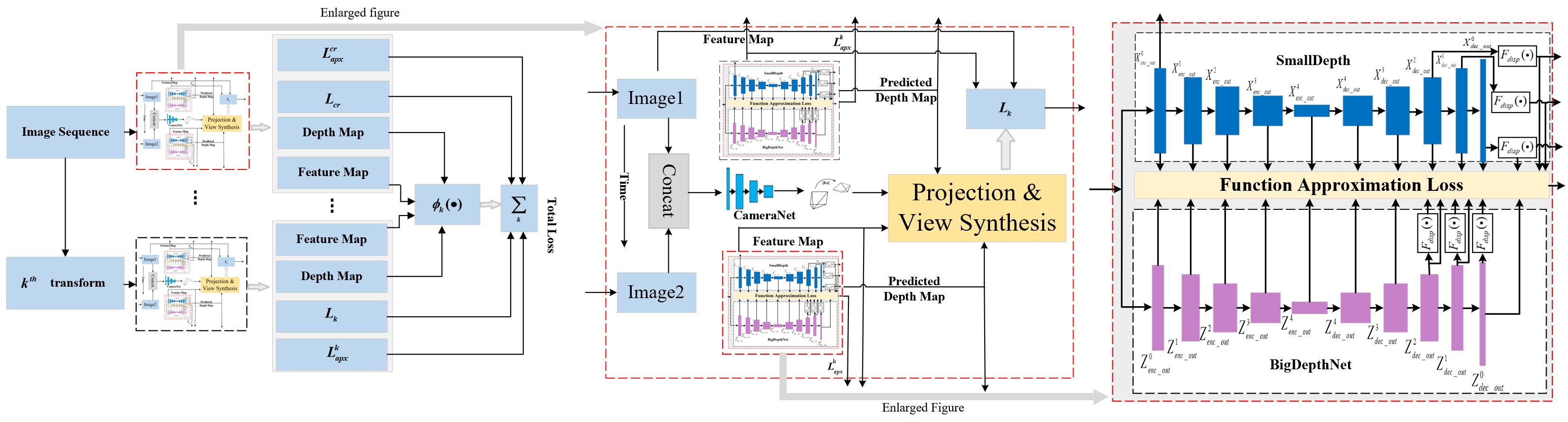}
	\caption{ Overview of the training architecture.}\label{fig:overview}	
	\vspace{-15pt}
\end{figure*}

\subsection{SmallDepth Design}\label{sec:smalldepthnet}

Given a convolutional layer, its complexity is shown in formula \eqref{eq_complexity}, where $C_i$ and $C_o$ are the numbers of filters of the input and output channels, $H_i,W_i,H_o,W_o$ are the height and width of the input and output feature maps, $K_h,K_w$  indicate the height and width of the filters, and $g$ is the number of groups. 
\begin{subequations}
	\label{eq_complexity}
	\begin{equation}
	\begin{aligned}
	\label{eq_complex_param}
	\resizebox{0.4\hsize}{!}{$Param=\frac{C_{i}*C_{o}*K_{h}*K_{w}}{g} $}
	\end{aligned}
	\end{equation}	
	\begin{equation}
	\begin{aligned}
	\label{eq_complex_flops}
	\resizebox{0.5\hsize}{!}{$FLOPs=H_{o}*W_{o}*\frac{C_{i}*C_{o}*K_{h}*K_{w}}{g} $}
	\end{aligned}
	\end{equation}	 
	\begin{equation}	
	\begin{aligned}	
	\label{eq_complex_memory_access_cost}	
\resizebox{0.88\hsize}{!}{$MAC=(H_{i}*W_{i}*C_i+K_h*K_w*\frac{C_i*C_o}{g}+H_o*W_o*C_o )*4 $}
	\end{aligned}
	\end{equation}  
\end{subequations}

According to formula \eqref{eq_complexity},  channel sparse connections, using a smaller kernel and reducing network width can help reduce model complexity. However, channel sparse connections hinder the information interaction between different channels,  reducing the filter size is not conducive to modeling pixel context information, and reducing the network width may reduce the diversity of features. To better balance the accuracy and complexity, we designed SmallDepth based on sparsity.

\paragraph{Sparse Downsampling}\label{sec:sd}

In contrast to the standard downsampling operation used in most DepthNets \cite{godard2019digging,wang2022cbwloss,bian2019unsupervised,zhou2017unsupervised,ranjan2019competitive}, where both the pixel context information and the dependencies between different channels are modeled simultaneously, we adopted a divide-and-conquer strategy in which these two kinds of information were encoded separately and then fused before being transmitted to the next layer, as shown in formula \eqref{eq_complex_spd}. The connections between different channels were removed, while the pixel context information was extracted by utilizing $F_{context}(\cdot)$. Only pixels at intervals of a certain distance in each channel can be connected during the channel interaction by utilizing $F_c(\cdot)$, resulting in parameter and FLOPs being reduced by $\frac{1}{K_h*K_w}+\frac{1}{g}$ times compared with the standard stride downsampling operation.
\begin{equation}
\label{eq_complex_spd}
\begin{aligned}
X_o=F_{context}(X_i)+ F_d(F_{c}(X_i))
\end{aligned}
\end{equation} 
where $F_{context}(\cdot)$/$F_c(\cdot)$represents filters whose kernel size and number of groups are greater than/equal to one,  $F_d(\cdot)$ is the drop function with a drop rate $pb_{sd}$ in formula \eqref{eq_random_drop}.

\paragraph{Double-scale Sparse Residual Module}\label{sec:dsr}

Although various DepthNets (e.g., \cite{zhou2021self,he2022ra,wangfei_hqdec}) proposed in recent years have improved the accuracy of depth estimation, the actual inference speed is slower \cite{wangfei_hqdec} than that of algorithms \cite{godard2019digging}, which can be attributed to the use of efficient basic blocks \cite{he2016deep}. However, the corresponding parameters and computational complexity are relatively high. To design a more lightweight and faster model, we revisited the network architecture \cite{godard2019digging,he2016deep} and proposed the double-scale sparse residual module. Different from the network used in \cite{godard2019digging,howard2019searching}, where two dense filters are directly stacked to form a basic residual block \cite{godard2019digging}, more blocks are directly stacked to increase the model's receptive field and feature extraction capability \cite{godard2019digging,howard2019searching}, and the parameters of the model are controlled by reducing the width of the network \cite{howard2019searching}. We design the feature encoding module by increasing the width and reducing the depth strategy to take full advantage of the parallel computing power of the GPU.

Specifically, given feature map $X_{i}\in \Re^ {C_i\times H_i \times W_i}$, we first transform it into a high-dimensional space by a learned function $F_{ct}(\cdot)$ with a $1\times 1$ filter according to formula \eqref{eq_spres_mid}. On the one hand, we model the local context information in each channel by a sparse channel function $F_{sc}(\cdot)$, obtained by eliminating the connection between the channels of the square filter, as shown in formula \eqref{eq_spres_sc}. Compared with dense filters, the number of parameters and FLOPs can be reduced by a factor of $g$ . On the other hand, to increase the receptive field without significantly increasing the complexity, we utilize sparse channel filters, which have the same filter shape and complexity (e.g., parameters, FLOPs, MAC) as $F_{sc}$, to model more distant information than $F_{sc}(\cdot)$ by establishing a relationship between the current pixel and non-adjacent surrounding pixels, as shown in formula \eqref{eq_spres_spc}. To reduce the loss of information caused by cutting off the connection between channels, the information contained in $X_{mid}$, where the pixel in each channel is related to the pixels in the other channels, was fused with $X_{sc}$ and $X_{spc}$, which were randomly dropped with a drop rate $pb_{dsr}$ to increase the robustness of the model during training, before channel transformation, as shown in formula \eqref{eq_spres_out}.
\begin{subequations}
	\label{eq_spres}
	\begin{equation}
	\label{eq_spres_mid}
	\begin{aligned}
	X_{mid}=F_{ct}(X_i) , X_{mid} \in \Re^{r*C_i \times H_i \times W_{i}}
	\end{aligned}
	\end{equation}
	\begin{equation}
	\label{eq_spres_sc}
	\begin{aligned}
	X_{sc}=F_{sc}(X_{mid}) ,    X_{sc} \in \Re^{r*C_i \times H_i \times W_{i}}
	\end{aligned}
	\end{equation}
	\begin{equation}
	\label{eq_spres_spc}
	\begin{aligned}
	X_{spc}=F_{spc}(X_{mid}) , X_{spc} \in \Re^{r*C_i \times H_i \times W_{i}}
	\end{aligned}
	\end{equation}
	\begin{equation}
	\label{eq_spres_out}
	\begin{aligned}
	X_{o}=F_{ct}(F_d(X_{sc})+F_d(X_{spc})+X_{mid})+X_{i}
	\end{aligned}
	\end{equation}	
\end{subequations}
\begin{equation}
\label{eq_random_drop}
\begin{aligned}
F_{d}(X)=\begin{cases}X, \qquad inference \\
\frac{B(1-pb)}{1-pb}*X , \qquad training	
\end{cases}
\end{aligned}
\end{equation}
\begin{equation} 
	\label{eq_drop_p}
	\begin{aligned}
	\resizebox{0.88\hsize}{!}{$pb=\begin{cases}
	(-\cos (\frac{\pi}{\lfloor N*r \rfloor})*i+1)*\frac{pb_{max}}{2}, 0\leq i \textless \lfloor N*r \rfloor \\
	(-\cos (\frac{\pi}{N-\lfloor N*r \rfloor}*i+\pi)+1)*\frac{pb_{max}}{2}, \lfloor N*r \rfloor \leq i \textless N
	\end{cases}$}
	\end{aligned}
\end{equation}	
where $B(\cdot)$, pb, $N_{iter}$ and $pb_{max}$ represent the Bernoulli sampling, drop probability at the batch dimension, total number of iterations during training, and maximum drop probability, respectively.

\paragraph{Sparse Upsample Module}\label{sec:sparse_upsample}%

In previous works \cite{godard2019digging,wang2022cbwloss}, dense square filters were used to fuse high-level information with low-level information and obtain the desired channel number before performing the interpolation operation and then refine the interpolated feature map. According to formula  \eqref{eq_complexity}, complexity is positively correlated with filter size and negatively correlated with the number of groups. Under the conditions of a certain number of parameters, number of groups, and  filter size, the MAC has a minimum value at $C_i=C_o$, as shown in formula \eqref{eq_mac_min}, according to the mean inequality. Therefore, we first transform the summed feature maps into the desired channel number by utilizing the smallest filter before focusing on the pixel context information by employing $F_{sc}(\cdot)$ with $C_i=C_o$. Furthermore, the interpolated feature maps  were first connected between channels by $F_{chs}(\cdot)$ (e.g., $1\times 1$ filters with $C_i=C_o$) before being refined by $F_{sc}(\cdot)$ with $C_i=C_o$.
\begin{equation}
	\label{eq_spres_up}
	\begin{aligned}
	X_o=F_{sc}( F_{chs}(F_{bi}(F_{sc}(F_{ct}(X_i)))))
	\end{aligned}
\end{equation} 
\begin{equation} 
\label{eq_mac_min}
\begin{aligned}
MAC \geq 4*(2*H*W*\sqrt{\frac{Param*g}{K_h*K_w}}+Param) 
\end{aligned}
\end{equation}
\paragraph{SmallDepth}          
To further reduce the inference time, instead of dense connections \cite{wangfei_hqdec} between the encoded and decoded feature maps, we only build connections, as shown in formula \eqref{eq_smalldepth}, between the encoded and the corresponding decoded feature maps, which have the same resolution as the encoded ones, to better balance speed and accuracy. 
\begin{equation}
\label{eq_smalldepth_enc}
\begin{aligned}
X_{enc}^{k}=\begin{cases}
F_s(X_{img})	, k=0 \\
F_{dsr}^{(k,n)}(F_{sd}(X_{enc}^{k-1})) , 1 \leq k \leq 4, 
\end{cases}
\end{aligned}
\end{equation}
\begin{equation}
n=\begin{cases}
1, 1\leq k \leq 2 \\
2, 2 \le k \leq 4
\end{cases}
\end{equation}
\begin{equation}
	\label{eq_smalldepth}
	\begin{aligned}
	X_{dec}^{k}=\begin{cases}
 X_{enc}^{k}+F_{up}(X_{dec}^{k+1})	, 0 \leq k \leq 2 \\
 X_{enc}^{k}+F_{up}(X_{enc}^{k+1}) , k=3
	\end{cases}
	\end{aligned}
\end{equation}
\begin{equation}
\label{eq_smalldepth_disp}
\begin{aligned}
X_{disp}^{k}=\frac{1}{2}\sum_{i=1}^{2}F_{i}(X_{dec}^{k})
\end{aligned}
\end{equation}
\begin{equation}
\label{eq_smalldepth_depth}
\begin{aligned}
D^{k}=1/(10*X_{disp}^{k}+0.01)
\end{aligned}
\end{equation}

where $X_{enc}^k$/$X_{dec}^{k}$ represents the encoded/decoded feature maps at stage $k$. $F_{up}(\cdot)$ is the sparse upsampling module in Sec. \ref{sec:sparse_upsample}. $n$ is the number of $F_{dsr}(\cdot)$ at stage $k$ in Sec. \ref{sec:dsr}. $F_{sd}(\cdot)$ is the sparse downsampling module in Sec. \ref{sec:sd}. $F_{i}(\cdot)$ represents the disparity regression module consisting of the filter with the spacing $i-1$ element between kernel elements followed by a sigmoid function.
\vspace{-10pt}
\subsection{Equivalent Transformation }\label{sec:eq_trans_module} 
Given pixel $x_{i,j}$ in the feature maps $X_{in}$, the corresponding output pixel with different scale information can be obtained  by utilizing the filters with different shapes to focus on the different context information according to formula \eqref{eq_eq_trans1}.
\begin{equation} 
\label{eq_eq_trans1}
\begin{aligned}
x_{i,j}^{o}=\sum_{t=1}^{T}\sum_{p_t,q_t,m_t,n_t}^{} w_{p_t,q_t}*x_{m_t,n_t}
\end{aligned}
\end{equation}
where {\small $m_t \in \{i-\lfloor  \frac{K_{h}^t}{2} \rfloor,..., i-1,i,i+1,...,i+\lfloor  \frac{K_{h}^t}{2} \rfloor \}$}, $n_t \in \{j-\lfloor  \frac{K_{w}^t}{2} \rfloor,..., j-1,j,j+1,...,j+\lfloor  \frac{K_{w}^t}{2} \rfloor \}$, $p_t \in \{-\lfloor  \frac{K_{h}^t}{2} \rfloor,..., -1,0,1,...,\lfloor  \frac{K_{h}^t}{2} \rfloor \}$, $q_t \in \{-\lfloor  \frac{K_{w}^t}{2} \rfloor,..., -1,0,1,...,\lfloor  \frac{K_{w}^t}{2} \rfloor \}$, and $w_{0,0}$ corresponding to $x_{i,j}$. $K_{h}^t$, $K_{w}^t$ denote the height and width of filter of the $t$-th shape, respectively. $T=\lceil \frac{K_h^T}{2}\rceil *\lceil \frac{K_w^T}{2}\rceil$ represents the total number of the filters with different shapes. $\lfloor \cdot \rfloor$/$\lceil \cdot \rceil$ represents rounding down/up. $K_h^T$/$K_w^T$ represents the maximum height/width among these filters.

The above strategy based on multi-shape filters, which can simultaneously fuse different context information, was usually used to improve the performance of the model \cite{ding2019acnet,szegedy2015going,mehta2018espnet}. However, compared with filters with a single shape, this structure has many disadvantages: a) The numbers of parameters, FLOPs, and MAC are larger. b) The inference speed is slower. c) The peak value of the memory occupation is larger. 

In order to enjoy both the performance gain brought by multi-shape filters and the advantages of a single-shape filter whose speed is faster and complexity is lower, we utilized multi-shape filters to model the different context information in parallel during training, while only a single-shape filter, whose height/width is the maximum height/width of the multi-shape filters and whose weights are equivalent weights of these multi-shape filters, was utilized to propagate the information flow during inference. 
To simulate more shapes of the filter, we proposed dropped convolution, named dropconv, where weights are dropped at the element-level, as shown in formula \eqref{eq_drop_conv}. 
\begin{equation}
\label{eq_drop_conv}
\begin{aligned}
F_{dw}(\mathbf{w})=\begin{cases}
\mathbf{w}, inference \; \mathbf{w} \in \Re^{C_o\times C_i \times K_h \times K_w} \\
\Upsilon \otimes \mathbf{w} ,  training \quad \Upsilon \in \Re^{C_o\times K_h \times K_w}
\end{cases}
\end{aligned}
\end{equation}
where $\otimes$ represents the element-level product. {\small  $C_i/C_o$} and {\small $K_h/K_w$} are the number of input/output channels and the height/width of the filters. $\Upsilon$ represents a random tensor obtained by Bernoulli sampling in $\Re^{C_o\times K_h \times K_w}$ space with a drop probability $pb_w= r_2*pb_t$.

Furthermore, different from formula \eqref{eq_eq_trans1}, during training, we randomly drop the weights of the filters of the $t$ th $(0\le t < T)$ branches  with $pb_t$ (each branch corresponds to the filter of one shape), as shown in formula \eqref{eq_eq_trans1_1}. The branches of t=0 correspond to identity mapping, which can be viewed as a constant filter with a weight of one for the central position and zero for the rest. The filters with  $K_{h}^{T}$ and  $K_{w}^{T}$ were utilized at the $T$ th and $T+1$ th branches while the filter of the $T$ th branch was dropped at both the branch-level with $r_1*pb_{t}$ according to formula \eqref{eq_random_drop} and the element-level with $r_2*pb_{t}$ according to \eqref{eq_drop_conv} and the filter of the $T+1$ th branch was used as the standard filter. For convenience of description, the product of the weight matrix and the corresponding region in the feature maps is defined as shown in formula \eqref{eq_dot_define}.
\begin{subequations}
\begin{equation}
\label{eq_eq_trans1_1}
\begin{aligned}
\resizebox{0.85\hsize}{!}{$ x_{i,j}^o= \lambda_{T+1} * \mathbf{w_{T+1}}\odot \mathbf{x_{T+1}} + \sum_{t=0}^{T} (F_d^{t}(\lambda_t * F_{dw}^t( \mathbf{w_t}))\odot \mathbf{x_t}) $}
\end{aligned}
\end{equation}
\begin{equation}
\begin{aligned}
\resizebox{0.55\hsize}{!}{$ \lambda_t=p_t/(\sqrt{var(\mathbf{x_t})+10^{-5}})$}
\end{aligned}
\end{equation}
\begin{equation}
\label{eq_dot_define}
\begin{aligned}
\mathbf{w_t}\odot \mathbf{x_t}=\sum_{p_t,q_t,m_t,n_t} w_{p_t,q_t}*x_{m_t,n_t}, 0\leq t \leq T+1
\end{aligned}
\end{equation}
\begin{equation}
\begin{aligned}
F_{dw}^t(\mathbf{w})=\begin{cases}
\mathbf{w}, 0\le t < T \\
F_{dw}(\mathbf{w}), t=T
\end{cases}
\end{aligned}
\end{equation}
\end{subequations}
where $p_t$ and $var(\mathbf{x_t})$ represent the learned parameter and the estimated variance. $F_d^t(\cdot)$ represents that the filter's weight on the t-branch was randomly dropped by $F_d(\cdot)$ in formula \eqref{eq_random_drop}.

Based on the addition criterion and multiplicative distribution law, formula \eqref{eq_eq_trans1_1} can be rewritten as formula \eqref{eq_eq_trans1_2}. As a result, not only is the diversity of information extracted by different filters  guaranteed because the weight of each filter is updated in a different direction, as shown in formula \eqref{eq_eq_trans_update_grad}, but also the training cost is reduced because the size of the feature map is much larger than the size of the filter. 
\vspace{-5pt}
\begin{equation}
\label{eq_eq_trans1_2}
\begin{aligned}
x_{i,j}^o= (\lambda_{T+1} * \mathbf{w_{T+1}}+ \sum_{t=0}^{T} (F_d^{t}(\lambda_t * F_{dw}^t(\mathbf{\hat {w}_t})))\odot \mathbf{x_T}
\end{aligned}
\end{equation}
\begin{equation}
\label{eq_eq_trans_update_grad}
\begin{aligned}
\mathbf{\hat w_t}  \leftarrow \begin{cases}
 \mathbf{\hat w_t} - \eta * \lambda_t *\frac{B(1-pb)}{1-pb}* \mathbf{ x_T}, 0\le t<T\\
 \mathbf{\hat w_t} - \eta * \lambda_t *\frac{B(1-pb)}{1-pb}*\Upsilon \otimes \mathbf{ x_T}, t=T \\
 \mathbf{\hat w_t} - \eta * \lambda_t *\mathbf{ x_T} , t=T+1
\end{cases}
\end{aligned}
\end{equation}
where $\eta$ denotes the learning rate. $\mathbf{\hat w_t}$ denotes the weights obtained by aligning $\mathbf{w_t}$ with $\mathbf{w_{T+1}}$ by filling in zeros.

Once the training is ocomplete, the corresponding equivalent weights $ \mathbf{w_{eq}}$ can be obtained according to formula \eqref{eq_eq_trans_weight} from the learned weights $\mathbf{\hat w_{t}}$ and $\lambda_t$. During inference, only the maximum filter in multi-shape filters was used, as shown in formula \eqref{eq_eq_trans1_infer}, resulting in faster speed and lower complexity than multi-shape filters because information  was propagated forward only once in a single-shape filter.
\vspace{-10pt}
\begin{equation}
\label{eq_eq_trans_weight}
\begin{aligned}
\mathbf{w_{eq}}=\lambda_{T+1} * \mathbf{w}_{T+1}+ \sum_{t=0}^{T} \lambda_t \mathbf{\hat w_{t}}
\end{aligned}
\end{equation}
\begin{equation}
\label{eq_eq_trans1_infer}
\begin{aligned}
x_{i,j}^o=\mathbf{w_{eq}} \odot \mathbf{x_T}
\end{aligned}
\end{equation}
\vspace{-10pt}
\subsection{Pyramid Loss}\label{sec:pyr_loss}
Given a network that is already designed and an input image $I_{cr} \in \Re ^{C \times H_{cr} \times W_{cr}}$, the perceptual capabilities of each layer for $I_{cr}$ are fixed. Compared with the current resolution image $I_{cr}$, longer range context information can be perceived by the same filters at the same position from the low-resolution image $I_{lr} \in \Re ^{C \times H_{lr} \times W_{lr}}$. On the other hand, the fine-grained information can be focused by the same filters at the same network depth from the high-resolution image $I_{hr} \in \Re ^{C \times H_{hr} \times W_{hr}}$. Second, the existing methods only utilize a signal (either the photometric differences between the source image and the corresponding synthesized image in current clips or the differences between the flipped versions) to guide the updating of model weights during each iteration, resulting in tighter constraints that minimize the photometric differences obtained by these two schemes in parallel being not utilized to guide the updating of model weights. Furthermore, we found that the depth $D_{cr}=F_{D}(I_{cr}|W_D)$ and the flipped depth $\hat D_{flip}$, obtained by flipping the depth $D_{flip}=F_D(I_{flip}|W_D)$, are not completely consistent, whereas the depths obtained by utilizing the above method should theoretically be the same. Third, similar to the flip transformation, although a color enhancement strategy  was also utilized to improve the robustness of the model to illumination changes in previous works \cite{godard2019digging,wangfei_hqdec}, the photometric differences,  are generated either from the source image and the corresponding synthesized image or from the color enhanced image and the corresponding synthesized image, resulting in only a kind of the photometric differences being utilized to guide the updating of weight during each iteration. In addition, similar phenomena can be found that the inconsistency still exists between the depth $D_{cr}$ and the color enhanced depth $D_{color}=F_{D}(I_{color}|W_D)$. To this end, we propose the pyramid loss shown in formula \eqref{eq_eq_pyloss}. 
\begin{equation}
\label{eq_eq_pyloss}\small 
\begin{aligned}
L_{py}=\sum_{k} (\alpha_k *L_{k}   + \beta_k * \phi_k(d_{cr},\hat {d}_k) + \gamma_k * \phi_k(f_{cr},\hat {f}_k) ) 
\end{aligned}
\end{equation}
\begin{equation}
\label{eq_eq_curr}
\begin{aligned}
L_{k}=\lambda_p * L_{p}^{k} +\lambda_d * L_d^{k} + \lambda_f * L_{feat}^{k} +L_{s}^{k}
\end{aligned}
\end{equation}
\begin{equation}\small 
\label{eq_phi} 
\begin{aligned}
\phi_k (a,b)=\begin{cases}
\|a-b \|	, k \in \{ color \} \\
 \frac{\sum \psi(a)*log^{\frac{\psi(a)}{\psi(b)}}+ \sum \psi(b)*log^{\frac{\psi(b)}{\psi(a)}}}{2}, k \in \{flip ,lr,hr\} \\
 0, k \in \{cr\} 
\end{cases}
\end{aligned} 
\end{equation} 
where $L_{p}^{k}$, $L_d^{k}$, $L_{feat}^k$, $L_s^{k}$, $k \in \{cr,lr,hr,flip, color\}$, are the bidirectional weighted photometric loss, the bidirectional depth structure consistency loss, the bidirectional feature perception loss, the smoothness loss \cite{wangfei_hqdec,wang2022cbwloss}, at the k-level. $\psi(\cdot)$ denotes the softmax function.  $\lambda_p, \lambda_d, \lambda_f$ are set  as in \cite{wangfei_hqdec,wang2022cbwloss}.

Compared with previous works \cite{wangfei_hqdec,wang2022cbwloss}, where only one-level loss $L_{cr}$ was used as a supervisory signal to guide the updating of the weights of DepthNet and PoseNet, the proposed pyramid loss can be used to improve the accuracy and robustness of the model by simultaneously using the signals calculated from different levels of input images and by enforcing the consistency between $D_{cr}$/$f_{cr}$ and $D_{k}/\hat {f}_{k}, k\in \{lr, hr, flip, color \}$. Specifically, the supervision signal calculated based on $I_{lr}$ allows the model to infer the depth information of the current pixel using the context information from a longer distance, but ignores the detailed information. The supervision signal calculated based on $I_{hr}$ allows the model to infer the depth information of the current pixel using more local details. The constraints, obtained by simultaneously minimizing the loss $L_{cr}$, $L_{flip}$ and the inconsistency between $d_{cr}/f_{cr}$ and $\hat {d}_{flip}/ \hat {f}_{flip}$, can improve the robustness of the model by ensuring  consistency of the scene depth at the same location because whether the same image is flipped or not, the scene depth at the same location should be the same. Similarly, the constraints, obtained by simultaneously minimizing the $L_{cr}$, $L_{color}$ and the inconsistency between $d_{cr}/f_{cr}$ and $\hat {d}_{color}/ \hat {f}_{color}$, can improve the robustness to illumination changes.

To reduce the usage of GPUs, instead of calculating the gradients after all the losses are summed, we calculate the gradients immediately after each level loss is obtained, and utilize the summed gradients to update the weights.

\vspace{-10pt}
\subsection{HQDecv2 }\label{sec:hqdecv2_method}
Although a high-quality decoder (HQDec) \cite{wangfei_hqdec} has achieved high accuracy in monocular depth estimation, the estimated disparity maps contain grid artifacts and its speed is relatively slow \cite{wangfei_hqdec}. To this end, we optimized the architecture of HQDec.

\begin{figure}[htbp]
 	\centering 
 	\vspace{-5pt}
 	\resizebox{0.8\linewidth}{!}{	
 	\includegraphics[height=1.8cm,width=3cm]{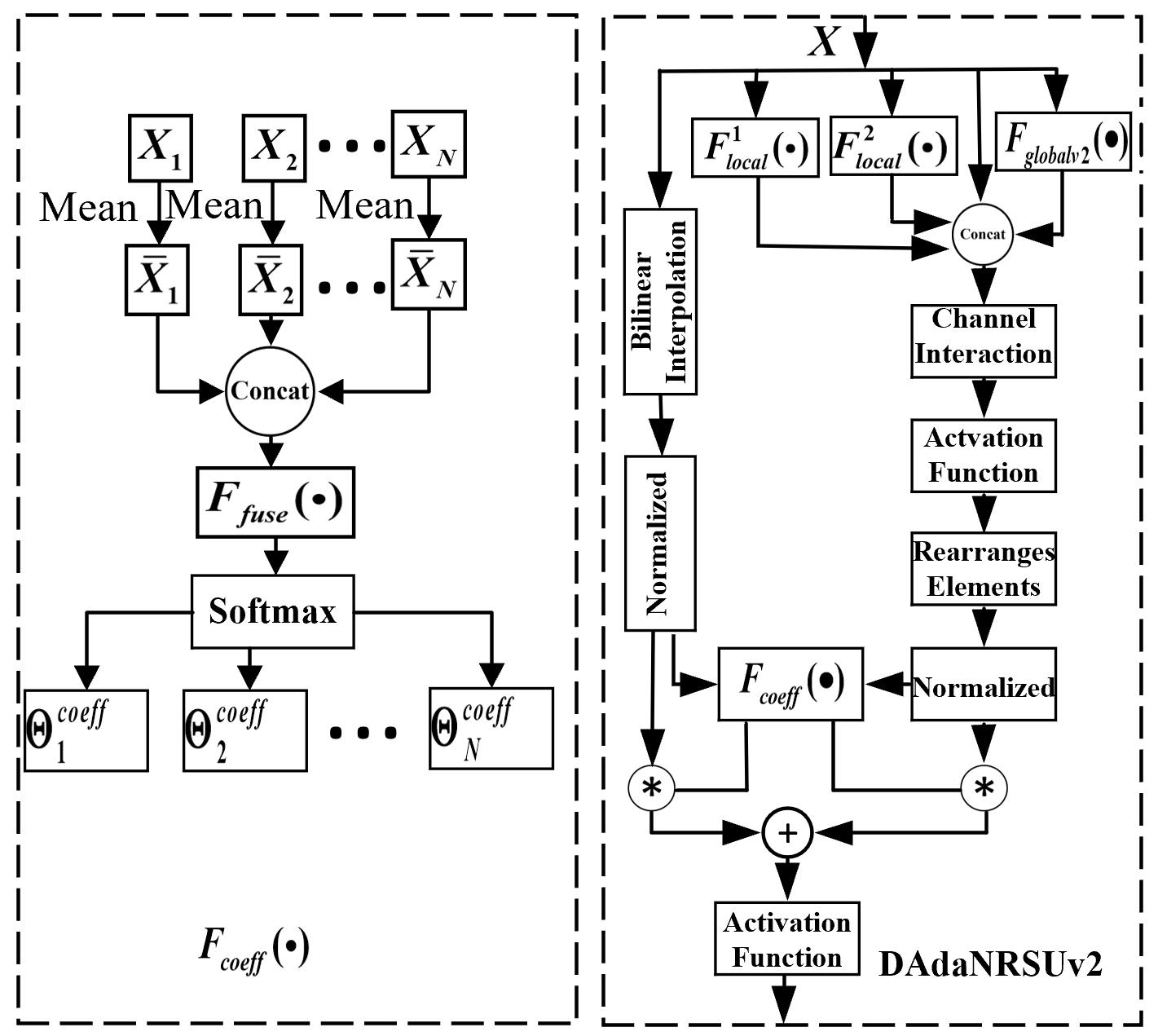}} 
 	\caption{AdaCoeff and DAdaNRSUv2 modules.}\label{fig:coefficient_modeling_upsample}	
 	\vspace{-10pt}
 \end{figure}

\paragraph{AdaCoeff}\label{sec:adacoeff}
It was found that the number of parameters of the learnable parameter tensor account for a large proportion in HQDec \cite{wangfei_hqdec} after analysis. To further reduce the complexity of HQDec, we propose an adaptive coefficient calculation module (AdaCoeff), as shown in Fig. \ref{fig:coefficient_modeling_upsample}, and use it to replace the learnable parameter tensors of each module (e.g., AdaIE, AdaRM, AdaAxialNPCAS, DAdaNRSU) used in HQDec \cite{wangfei_hqdec}.

Given the feature map $X_{i}$, $1 \le i \le n $ of n branches, the adaptive coefficient can be obtained according to formula \eqref{eq_coeff}. Concretely, we first calculate the mean $\bar {X_i}$ of $X_i$ and concatenate them at the channel dimension via the function $F_{cat}(\cdot)$. The relations between branches can be modeled by $F_{fuse}(\cdot)$ which can be implemented by stacking multilayer square filters. To increase the fitting ability of $F_{fuse}(\cdot)$, we first expand the concatenated feature map into a high-dimensional space, and then model the dependencies between branches using a square filter. To obtain the corresponding adaptive coefficient matrix, we squeezed the processed feature map into the original feature space using a square filter and calculated the probability values in different channels but in the same spatial position using softmax. Finally, the adaptive coefficient corresponding to each branch was obtained by slicing the obtained adaptive coefficient matrix at the channel dimension by utilizing $F_{split}(\cdot)$.
\begin{equation}\small 
\label{eq_coeff}
\begin{aligned}
& \Theta_{1}^{coeff},...,\Theta_{n}^{coeff}=F_{coeff}(\bar X_1,...,\bar X_n) \\
& =F_{split}(Softmax(F_{fuse}(F_{cat}(\bar X_1,...,\bar X_n)))
\end{aligned}
\end{equation}

\paragraph {AdaRMv2}\label{sec:adarmv2}
To address grid artifacts presented in the estimated disparity maps, we reconceptualized HQDec \cite{wangfei_hqdec}.  After the analysis, we found that the grid artifacts could be derived from two aspects: (1) the original feature maps were directly divided into non-overlapping sub-feature maps to calculate the global feature maps. (2) transposed convolution was used to map the computed global feature vector to the original feature map space. The reasons are as follows: (1) although we can calculate the global correlation of these sub-feature maps by dividing them into non-overlapping sub-feature maps, it is easy to ignore the boundary correlation between sub-feature maps. (2) in the process of mapping vectors to feature maps using transposed convolution, zero filling is performed around each pixel before performing convolution, resulting in the amplification of pixel differences.

\begin{figure}[htbp]
	\centering 
	\vspace{-10pt}	
	\resizebox{\linewidth}{!}{
		\includegraphics[scale=0.35]{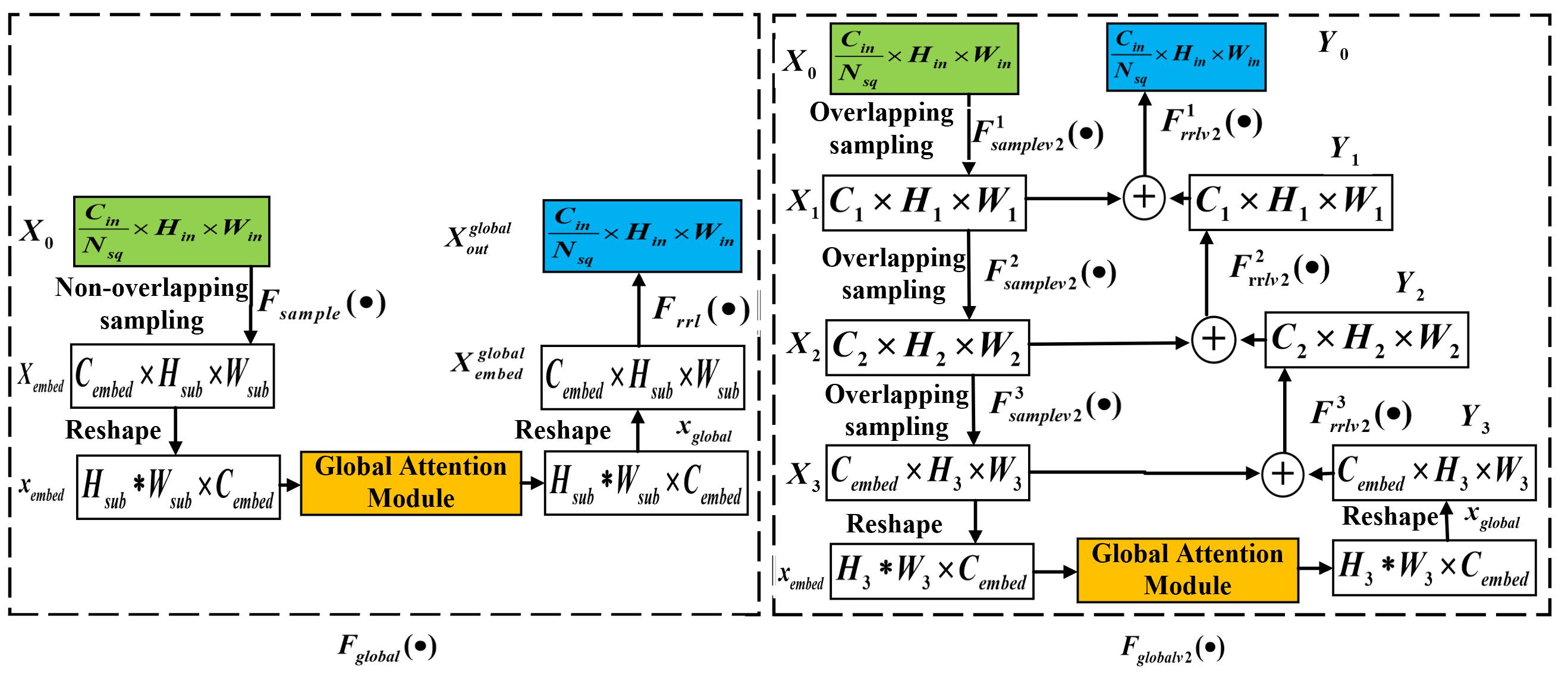}	}
	\caption{Global feature maps calculation module.  }\label{fig:global_modeling}	
	\vspace{-15pt}
\end{figure}

To this end, we (1) utilized multilevel overlapping sampling to divide the original feature map into the desired sub-feature maps; (2) mapped the global feature vector to the coarse-grained feature map by interpolation to reduce the difference between adjacent pixels, and used a square filter to refine the coarse-grained feature map; (3) utilized skip connections to reduce boundary differences, as shown in Fig. \ref{fig:global_modeling}. 

Concretely, given feature maps $X_{in}$, feature maps $X_{local}$ with local context information can be obtained according to formula \eqref{eq_local}, where $N_{sq}$ represents the squeezing ratio. 
\begin{subequations}
	\label{eq_local}
	\begin{equation}\label{eq_local2}	
	X_{0}=F_{squeeze}(X_{in}),X_0\in \Re^{\frac{C}{N_{sq}}\times H_{in} \times W_{in}}	
	\end{equation}
	\begin{equation}	
	X_{mid}=F_{sc}(X_{0}),X_{mid}\in \Re^{\frac{C}{N_{sq}}\times H_{in} \times W_{in}}	
	\end{equation}
	\begin{equation}	
	X_{local}=F_{unsqueeze}(X_{mid}),X_{local}\in \Re ^{C\times H_{in} \times W_{in}}	
	\end{equation}	   
\end{subequations}

We first divide $X_{0}$ into overlapping sub-feature maps by utilizing $F_{samplev2}(\cdot)$ and then embed it into feature vectors $x_{embed} \in \Re^{H_{sub}*W_{sub} \times C_{embed}}$, as shown in Fig. \ref{fig:global_modeling}. The multi-head attention (MHA) mechanism \cite{wangfei_hqdec} was utilized to calculate the corresponding global feature vectors $x_{global} \in \Re ^{H_{sub}*W_{sub} \times C_{embed}}$ according to previous work \cite{wangfei_hqdec}. To further alleviate the grid phenomenon in the estimated disparity map, we replaced the previous resolution recovery function $F_{rrl}(\cdot)$, which was implemented based on dense transpose convolution filters and was used to map the global feature vector $x_{global}$ to the original feature map space, in \cite{wangfei_hqdec} with the improved version $F_{rrlv2}(\cdot)$,  which can be implemented by interpolation followed with square filter. Instead of 
zero-fill around each pixel in the low-resolution feature map, we use the information obtained by bilinear interpolation to fill around each element, and then fine-tune the boundary by utilizing a local filter. Moreover, a skip connection was utilized to propagate the boundary information, contained in the pre-embedding feature map, into the recovered feature map, as shown in Fig. \ref{fig:global_modeling}. 
\begin{subequations}
\begin{equation}
\begin{aligned}
X_i=F_{samplev2}^{i}(X_{i-1}), X_i \in \Re^{C_i\times H_i \times W_i}
\end{aligned}
\end{equation}
\begin{equation}
\begin{aligned}
F_{samplev2}^{i}(X)=F_{ct}(F_{sc}^{i}(X))
\end{aligned}
\end{equation}
\begin{equation}
\begin{aligned}
Y_{i-1}=F_{rrlv2}^{i}(X_i+Y_i)
\end{aligned}
\end{equation}
\begin{equation}
\begin{aligned}
X_{global}^{out}= F_{unsqueeze}(Y_0)
\end{aligned}
\end{equation}
\begin{equation}
\begin{aligned}
\resizebox{0.85\hsize}{!}{$ X_{adarmv2} =\Theta_{identity}^{coeff}*X_{in} + \Theta_{local}^{coeff}*X_{local}  + \Theta_{global}^{coeff}*X_{global}^{out}$}
\end{aligned}
\end{equation}
\begin{equation} 
\begin{aligned}
\resizebox{0.85\hsize}{!}{$ \Theta_{identity}^{coeff},\Theta_{local}^{coeff},\Theta_{global}^{coeff}=F_{coeff}(\bar X_{in},\bar X_{local},\bar X_{global}^{out}) $}
\end{aligned}
\end{equation}
\end{subequations}
where $i\in  {1,2,3}$, $C_3=C_{embed}$. $F_{sc}^{i}(\cdot)$ represents a channel sparse filter with kernel size $k_i$ and stride $s_i$.
\begin{equation}
\begin{aligned}
(s_i,k_i)=\begin{cases}(4,5), min(H_{i-1},W_{i-1})>16 \\
(2,3), 3< min(H_{i-1},W_{i-1})\le 16 \\
(1,3), other 
\end{cases}
\end{aligned}
\end{equation}

\paragraph {DAdaNRSUv2}

Instead of the DAdaNRSU used in HQDec \cite{wangfei_hqdec}, we utilized DAdaNRSUv2 in Fig. \ref{fig:coefficient_modeling_upsample} 
as the upsampling module of HQDecv2 to address the grid phenomenon and reduce complexity because AdaRM \cite{wangfei_hqdec}, which could lead to grid phenomenon (see Sec. \ref{sec:adarmv2}),  was also used to model local and global information in DAdaNRSU.

Specifically, (a) instead of the scheme in which the filter is directly used to expand the number of feature maps by a factor of four to recover high-resolution feature maps \cite{wangfei_hqdec}, the number of feature maps was expanded according to formula \eqref{eq_expand_chs_num} where we ensure that each function (e.g., $F_{local}^1(\cdot)$, $F_{local}^2(\cdot)$, $F_{global}(\cdot)$) has an equal number of input and output channels to minimize the complexity (e.g., MAC) based on the analysis in Sec. \ref{sec:smalldepthnet}. Compared with the previous scheme \cite{wangfei_hqdec}, another advantage is that this approach can aggregate and exchange information at different scales simultaneously. (e.g., the identity mapping ensures that the information flow of the previous stage is propagated to the next stage. $F_{local}^1(\cdot)$ and $F_{local}^2$ can model local context information at different scales. $F_{globalv2}(\cdot)$ not only models the global context information but also eliminates the grid phenomenon.) The information at different scales can be exchanged by the channel interaction function $F_{ci}(\cdot)$ used in AdaAxialNPCAS \cite{wangfei_hqdec}. (b) the learnable parameter tensor in DAdaNRSU \cite{wangfei_hqdec}, which was used to adaptively fuse high-resolution feature maps obtained by bilinear interpolation and upsampling schemes based on rearranging elements,  was replaced with AdaCoeff shown in Fig. \ref{fig:coefficient_modeling_upsample}.
\begin{subequations}\label{eq_upsample}
\begin{equation}
\begin{aligned}
X_o=\Theta_{ip}*F_{ip}(X)+ \Theta_{pixel}*F_{pixelshuffle}(X_{ci})
\end{aligned}
\end{equation}
\begin{equation}\small \label{eq_expand_chs_num}
\begin{aligned}
X_{ci}=F_{ci}(F_{cat}([F_{local}^1(X),F_{local}^2(X),F_{globalv2}(X),X])
\end{aligned}
\end{equation}
\end{subequations}
where $\Theta_{ip}$ and $\Theta_{pixel}$ represent  the corresponding adaptive coefficients obtained by formula \eqref{eq_coeff}. $F_{cat}(\cdot)$ represents the concatenation function along the channel dimension. $F_{local}^1(\cdot)$ and $F_{local}^2(\cdot)$ represent $3\times3$ filters and $5\times5$ filters, respectively.


\paragraph {AttDispv2}

Similar to AdaRMv2 and DadaNRSUv2, the module, which was used to model the global context information in the previous AttDisp \cite{wangfei_hqdec}, was replaced with $F_{globalv2}(\cdot)$ shown in Fig. \ref{fig:global_modeling}.

\subsection{Function Approximation Loss}\label{sec:func_approx_loss_method}  %

Although the accuracy of the depth directly recovered from monocular video in terms of being fully unsupervised greatly improved, more complex network models are often needed to fit the data distribution characteristics in the input images. However, complex network models tend to increase the time complexity and space complexity of algorithms, resulting in a slower inference speed and greater complexity. On the other hand, although the small model has the advantage of faster inference speed and lower complexity, its accuracy is lower than that of the large model.

\begin{figure}[htbp]
	\centering
	\includegraphics[scale=0.18]{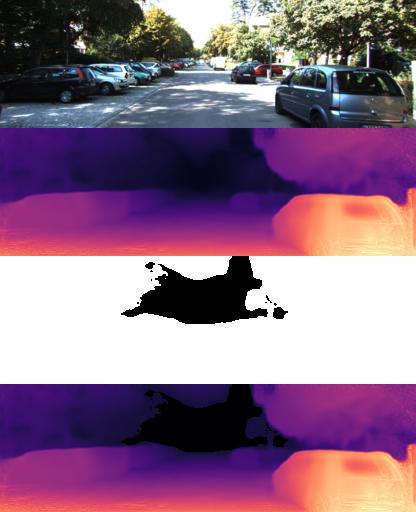}
	\includegraphics[scale=0.18]{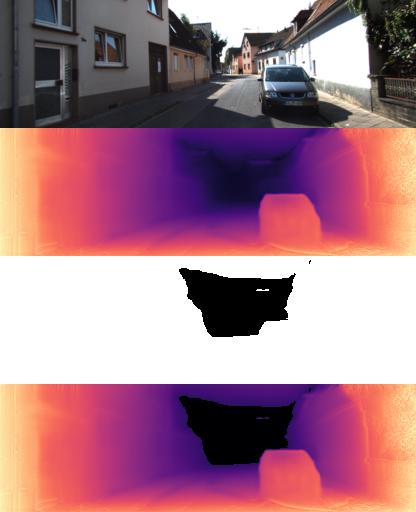}	
	\includegraphics[scale=0.06]{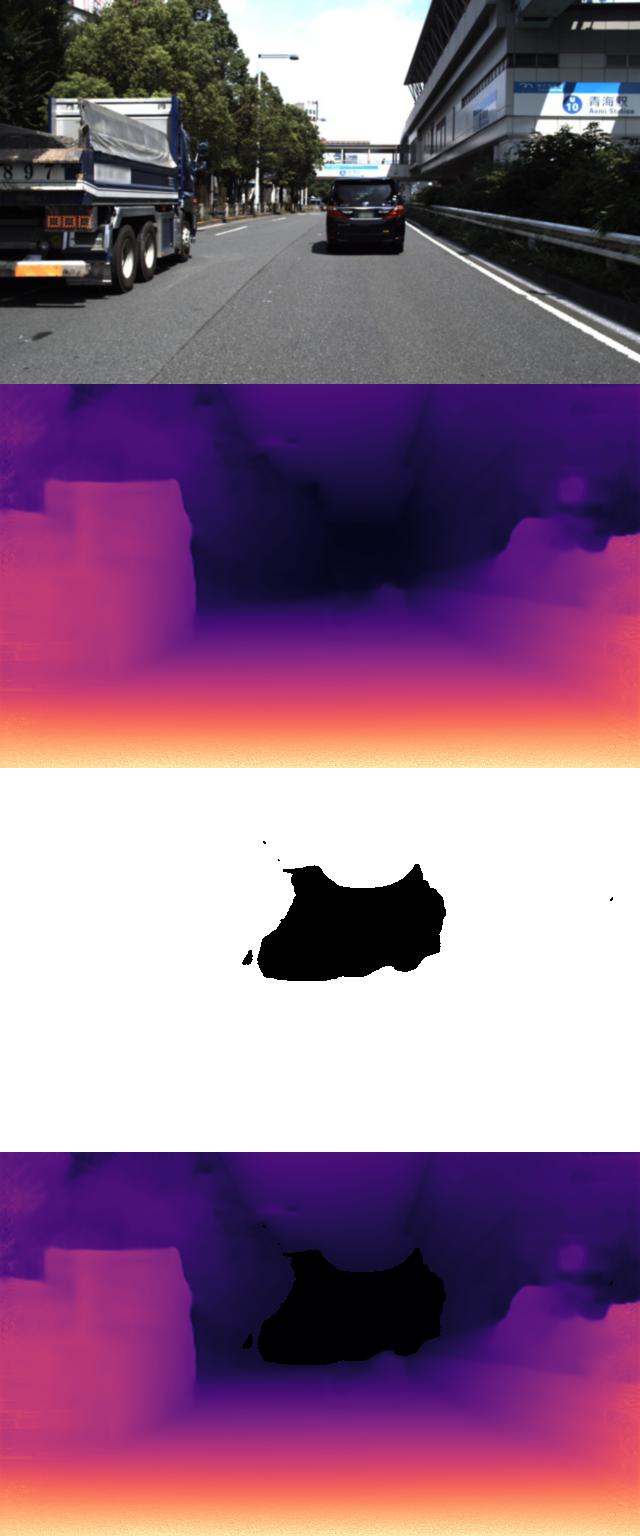}	
	\includegraphics[scale=0.06]{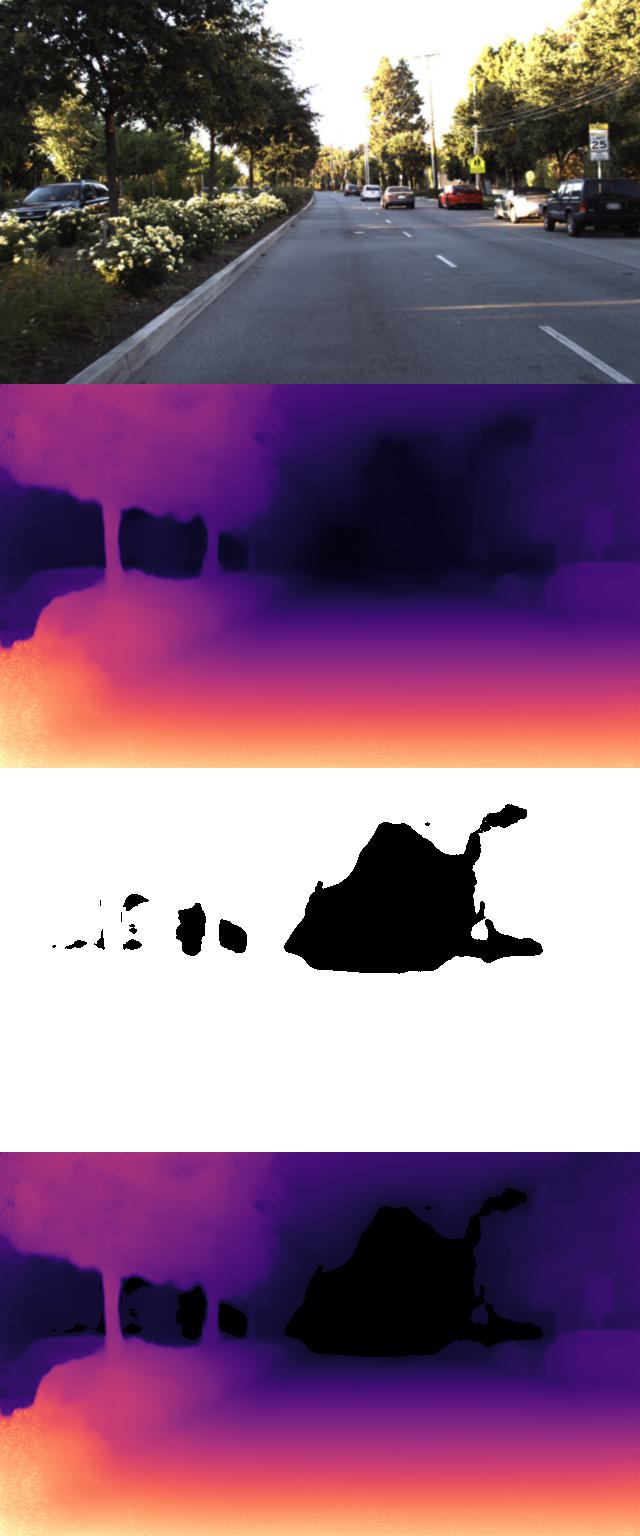}	
	\caption{The mask on KITTI and DDAD dataset.}
	\label{fig:mask_sample}
\end{figure}

In order to enjoy the advantages of the low complexity and fast inference speed of small models and the high prediction accuracy of large models, we propose the APX, as shown in formula \eqref{eq_function_approximation_loss_error}.  Different from previous works \cite{petrovai2022exploiting,sun2023sc,shi20233d,li2023sense,lyu2021hr}, we utilize the probability distribution difference between results (e.g., feature and disparity maps), predicted by small and large models, respectively, as a supervisory signal to monitor the updating of the weights of the small model, making its distribution characteristics similar to those of the large model.

Although the depth in most regions can be accurately predicted by a large model, there exists some regions where the depths may be inaccurately estimated. e.g., the regions containing moving objects and occlusions. The depth of these regions may be mapped to infinity. Moreover, the prediction was more accurate in the near regions than those in the far regions. The pseudo-depth in these regions provides false supervisory signals for lightweight models, resulting in incorrect depth predictions. Different from existing works \cite{lyu2021hr,petrovai2022exploiting,shi20233d,sun2023sc}, where the pseudo-depth, generated by the pretrained large model based on self-supervised learning was assumed to be correct by default \cite{lyu2021hr}, the inconsistent prediction depth map caused by the occluded area was filtered out \cite{petrovai2022exploiting}, the false prediction caused by the reflective surfaces was processed by using the ground pose \cite{shi20233d}, or the pretrained network based on ground truth was used to generate pseudo-depth \cite{sun2023sc}, we proposed a simple and practical position mask scheme, as shown in formula \eqref{eq_mask_apx}, to filter out inaccurate predictions, as shown in Fig. \ref{fig:mask_sample} .

\begin{subequations}
	\begin{equation}\label{eq_function_approximation_loss_error}
	\begin{aligned} 
\resizebox{0.85\hsize}{!}{$	L_{apx}= \sum_{j} \lambda_{j}\sum_{i=0}^4 P(Z_{j}^{i},X_{j}^{i})  + \lambda_{disp}\sum_{0}^3 P(M_i*Z_{disp}^{i},M_i*X_{disp}^{i}) $}
	\end{aligned}
	\end{equation}
	\begin{equation}
	\begin{aligned}\label{eq_mask_apx}
	\resizebox{0.5\hsize}{!}{$M_i=Z_{disp}^{i}> 0.3*median(Z_{disp}^{i}) $}
	\end{aligned}
	\end{equation}
	\begin{equation}\label{eq_pde_error}
	\begin{aligned}
	\resizebox{0.55\hsize}{!}{$P(a,b)=\frac{1}{2}\sum_{m}(\psi_m(a)*log^{\frac{\psi_m(a)}{\psi_m(b)}}) $}
	\end{aligned}
	\end{equation}
\end{subequations}
where  $X_{j}^{i}/Z_{j}^{i}$, $j\in \{enc,dec\}$ 
represent the feature maps output by the small/large model at stage $i$, respectively. $X_{disp}^{i}$/ $Z_{disp}^{i}$ represent the corresponding disparity at the $i$th scale obtained by the small/large model. $m \in \{h,w\}$, $m=h/w$ represents the computed probability distribution along the height/ width direction.

\vspace{-10pt}
\section{Experiments}\label{sec:experiments}
\subsection{Dataset}
We conducted experiments on the KITTI RAW dataset \cite{geiger2013vision}and the DDAD dataset \cite{guizilini20203d}. Similar to previous related works \cite{zhou2017unsupervised,ranjan2019competitive,godard2017unsupervised,wang2022cbwloss,wangfei_hqdec}, the KITTI RAW dataset was split as in \cite{eigen2014depth}, with approximately 40k frames used for training and 5k frames used for validation. We evaluated DepthNet on test data consisting of 697 test frames in accordance with Eigen's testing split. The DDAD dataset was split as in \cite{guizilini20203d}, where 150 scene videos from the front camera were used for training and 50 scene videos were used for evaluation purposes.
\vspace{-5pt}
\subsection{Experiment Details}
The proposed learning framework was implemented using the PyTorch Library \cite{paszke2017automatic}. The training set was augmented using ColorJitter by randomly changing the brightness, contrast, saturation, and hue of each image, as well as performing random scaling, random horizontal flipping, and random cropping. During the training process, three consecutive video frames were used as a training sample and fed to the model. If no special instructions, the loss function (namely $L_{p}^{cr}$ in Sec. \ref{sec:pyr_loss}) proposed in \cite{wang2022cbwloss} was used as a supervisory signal to jointly optimize DepthNet and CameraNet on an RTX 4090 GPU from scratch using the AdamW \cite{loshchilov2017decoupled} optimizer. ETM and $L_{py}$ were not utilized in HQDecv2 due to hardware limitations.

\subsection{Comparison with State-of-the-Art Methods}
In Tab. \ref{Tab:depth_compared_previous_method_80m},  we quantitatively compare the proposed SmallDepth and HQDecv2 with previous methods on KITTI dataset.  In Tab. \ref{Tab:param_time}, we compare the complexity of the different methods  under the same conditions. The results show that both the large model (HQDecv2) and the proposed small model(SmallDepth) are superior to the published methods. Although only a single frame of information is used during inference, the proposed method can still achieve similar or even better performance than the method using multiframe information during inference. Compared with the previous methods, the proposed SmallDepth has the fastest speed, the lowest complexity, and achieves higher prediction accuracy, as shown in Tab. \ref{Tab:depth_compared_previous_method_80m} and Tab. \ref{Tab:param_time}. The results evaluated on the more challenging DDAD dataset \cite{guizilini20203d} in Tab.  \ref{Tab:depth_compared_on_ddad} show that the proposed method outperformed all previously published self-supervised methods, including the approaches that utilize multiple frames to predict the corresponding depth and methods that use semantic labels as supervision signals.

\begin{table}[htbp]\small
	\vspace{-5pt}
	
	\setlength\tabcolsep{2pt}
	\centering
	\resizebox{0.95\linewidth}{!}{
		\begin{tabular}{llcccc} %
			\toprule %
			\multicolumn{1}{l}{}&\multicolumn{1}{l}{Method}& \multicolumn{1}{c}{Param({\footnotesize M})} &\multicolumn{1}{c}{GFLOPs({\footnotesize G})}& \multicolumn{1}{c}{FPS} & \multicolumn{1}{c}{GPU-Utili}\\

			\hline %
			\multirow{11}{*}{\makecell{Same Settings. \\(e.g., RTX 4090,\\  BatchSize=1,\\ $128\times 416$)} }
			&\multicolumn{1}{l}{Zhao et al.\cite{zhao2022monovit}}&27.87&13.02& 73 &64\%\\

			&\multicolumn{1}{l}{Zhou et al.\cite{zhou2021self}}&10.87&6.84& 69 &67\%\\
			
			&\multicolumn{1}{l}{He et al.\cite{he2022ra}}&9.98&4.67& 68 &65\%\\ 
			&\multicolumn{1}{l}{Guizilini et al.\cite{guizilini20203d}}&128.29&89.04& 28 &99\%\\

			&\multicolumn{1}{l}{Godard et al.\cite{godard2019digging}}&14.84&3.49& 212 &72\%\\
			
			
			&\multicolumn{1}{l}{Lyu et al.\cite{lyu2021hr}}&14.61&7.84& 172 &66\%\\
			
			
			&\multicolumn{1}{l}{Wang et al.\cite{wang2022cbwloss}}&32.52&7.22& 145&70\% \\ 
			
			&\multicolumn{1}{l}{Yan et al.\cite{yan2021channel}}&58.34&17.66& 129 &86\%\\ 
			
			&\multicolumn{1}{l}{HQDec \cite{wangfei_hqdec}}&29.29&5.90&36&51\%\\

			&\multicolumn{1}{l}{HQDecv2 (ours)}&24.65&3.90&30&40\%\\  
			&\multicolumn{1}{l}{SmallDepth (ours)}&2.35&0.42&557&8\%\\ 
			
			\bottomrule 
	\end{tabular}}
	\caption{ Complexity and offline inference time comparison.  }\label{Tab:param_time}
	\vspace{-10pt}
\end{table}

\begin{table*}[htbp]\small
	\setlength\tabcolsep{2pt}
	\centering	
	\resizebox{0.75\linewidth}{!}{
		\begin{tabular}{lcccccccccccccccc} %
			\toprule 
			
			
			
			

			{\footnotesize Method}&{\footnotesize DE}&{\footnotesize PE}&{\footnotesize Data}&{\footnotesize RES}&{\footnotesize Sup + Multi-Fr?}&\footnotesize AbsRel$\downarrow$&\footnotesize SqRel$\downarrow$&\footnotesize RMSE$\downarrow$&\footnotesize RMSE log$\downarrow$& \footnotesize $\delta_1$$\uparrow$ &\footnotesize $ \delta_2$$\uparrow$&\footnotesize $\delta_3$$\uparrow$&\\

			\midrule

			\rowcolor{blue!5} \multicolumn{13}{c}{KITTIRAW Eigen Test Set\cite{eigen2014depth}}\\
			
			\multicolumn{1}{l}{Watson et al. \cite{watson2021temporal}}&RN18&RN18&K+CS&192$\times$640&M+Multi-Fr&0.098&0.770&4.459&0.176& 0.900&0.965&0.983&\\


		     \multicolumn{1}{l}{Miao et al. \cite{miao2023ds}}&RN18&RN18&K&192$\times$640&M+Multi-Fr&0.095&0.698&4.329&\underline{0.173}& 0.905&0.966&\underline{0.984}&\\
		

			 \multicolumn{1}{l}{Feng et al. \cite{feng2023iterdepth}}&-&-&K&192$\times$640&M+Multi-Fr&\underline{0.093}&0.723&4.317&0.174& \textbf{0.908}&0.965&\underline{0.984}&\\

			\multicolumn{1}{l}{Guizilini et al. \cite{guizilini2022multi}}&Depthformer&RN18&K+CS&192$\times$640&M+Multi-Fr&\textbf{0.090}&\underline{0.661}&\underline{4.149}&0.175&0.905&\underline{0.967}&\underline{0.984}&\\

			\multicolumn{1}{l}{Zhou et al. \cite{zhou2022self}}&Swin&-&K&384$\times$1280&S&\textbf{0.090}&\textbf{0.538}&\textbf{3.896}&\textbf{0.169}&\underline{0.906}&\textbf{0.969}&\textbf{0.985}&\\
			
			\midrule  

			\multicolumn{1}{l}{Li et al. \cite{li2020unsupervised}}&DN&RN18&K&128$\times$416 &M&0.130&0.950&5.138&0.209 &0.843&0.948&0.978&\\		

			\multicolumn{1}{l}{Godard et al. \cite{godard2019digging}}&RN18&RN18&K&128$\times$416  &M&0.128&1.087&5.171&0.204& 0.855&0.953&0.978& \\

			\multicolumn{1}{l}{Yan et al. \cite{yan2021channel}}&RN50&RN50&K&128$\times$416 &M&{0.116}&{0.893}&{4.906}&{0.192} &{0.874}&{0.957}&{0.981}& \\

			\multicolumn{1}{l}{HQDec \cite{wangfei_hqdec}}&EffV2s&FBv3&K&128$\times$416  &M&{0.103}&{0.706}&{4.569}&{0.176}&{0.882}&{0.962}&\textbf{0.985}& \\
			
			\multicolumn{1}{l}{HQDec \cite{wangfei_hqdec}$^\ddagger$}&EffV2s&FBv3&K&128$\times$416  &M&{0.099}&{0.693}&{4.494}&\textbf{0.173}&{0.887}&\underline{0.963}&\textbf{0.985}& \\
			
			\multicolumn{1}{l}{\textbf{SmallDepth(Ours)}}&-&FBv3&K&128$\times$416  &M&0.097&\underline{0.674}&4.554&0.177& 0.888&0.961&\underline{0.983}& \\
			\multicolumn{1}{l}{\textbf{SmallDepth(Ours)$^\ddagger$}}&-&FBv3&K&128$\times$416  &M&\textbf{0.094}&\textbf{0.662}&4.494&\underline{0.174}& \underline{0.893}&0.962&\underline{0.983}& \\
			
			\multicolumn{1}{l}{\textbf{HQDecv2(Ours)}}&EffV2s&FBv3&K&128$\times$416  &M&0.098&0.695&\underline{4.472}&0.179& \underline{0.893}&\underline{0.963}&\underline{0.983}& \\
			\multicolumn{1}{l}{\textbf{HQDecv2(Ours)}$^\ddagger$}&EffV2s&FBv3&K&128$\times$416  &M&\underline{0.095}&0.682&\textbf{4.400}&0.176& \textbf{0.899}&\textbf{0.964}&\underline{0.983}& \\

			\midrule 

			
			\multicolumn{1}{l}{Wang et al. \cite{wang20223d}}&DispRN18&PN7&K&256$\times$832  &M&0.109&0.790&4.656&0.185& 0.882&0.962&0.983& \\

			\multicolumn{1}{l}{Lyu et al. \cite{lyu2021hr}}&RN18&RN18&K&192$\times$640  &M&0.109&0.792&4.632&0.185& 0.884&0.962&0.983& \\
			
			\multicolumn{1}{l}{Liu et al. \cite{liu2023self}}&RN50&RN18&K&192$\times$640  &M&0.106&0.718&4.520&0.180& 0.886&0.964&0.983& \\

			\multicolumn{1}{l}{Petrovai et al. \cite{petrovai2022exploiting}}&RN50&RN18&K+CS&192$\times$640  &M&{0.100}&{0.661}&{4.264}&{0.172} &{0.896}&\underline{0.967}&\textbf{0.985}& \\

			\multicolumn{1}{l}{Zhao et al. \cite{zhao2022monovit}}&MPVit&RN18&K&192$\times$640&M&0.099&0.708&4.372&0.175& {0.900}&\underline{0.967}&\underline{0.984}& \\
			
			
			\multicolumn{1}{l}{Hou et al. \cite{hou2024self}}&Uformer&RN18&K&192$\times$640&M&0.098&0.726&4.375&0.177& 0.891&0.967&0.986& \\

			\multicolumn{1}{l}{He et al. \cite{he2022ra}}&HR18&RN18&K&192$\times$640 &M&{0.096}&\textbf{0.632}&{4.216}&0.171& {0.903}&\textbf{0.968}&\textbf{0.985}& \\
			
			\multicolumn{1}{l}{Han et al. \cite{han2022transdssl}}&Swin&PN7&K&192$\times$640 &M&0.098&0.728&4.458&0.176& 0.898&0.966&\underline{0.984}& \\

			\multicolumn{1}{l}{HQDec \cite{wangfei_hqdec}}&EffV2s&FBv3&K&192$\times$640 &M&{0.096}&0.654&4.281&\underline{0.169}& 0.896&0.965&\textbf{0.985}& \\
			
			\multicolumn{1}{l}{HQDec \cite{wangfei_hqdec}$^\ddagger$}&EffV2s&FBv3&K&192$\times$640  &M&\underline{0.092}&\underline{0.642}&{4.233}&\textbf{0.167}& {0.899}&{0.966}&\textbf{0.985}& \\

			\multicolumn{1}{l}{\textbf{SmallDepth(Ours)}}&-&FBv3&K&192$\times$640  &M&0.097&0.670&4.550&0.178& 0.890&0.961&0.983& \\
			\multicolumn{1}{l}{\textbf{SmallDepth(Ours)$^\ddagger$}}&-&FBv3&K&192$\times$640  &M&0.094&0.660&4.497&0.175& 0.895&0.961&0.982& \\
			
			\multicolumn{1}{l}{\textbf{HQDecv2(Ours)}}&EffV2s&FBv3&K&192$\times$640  &M&\underline{0.092}&0.680&\underline{4.198}&\underline{0.169}& \underline{0.909}&\underline{0.967}&\underline{0.984}& \\
			\multicolumn{1}{l}{\textbf{HQDecv2(Ours)}$^\ddagger$}&EffV2s&FBv3&K&192$\times$640  &M&\textbf{0.087}&0.664&\textbf{4.140}&\textbf{0.167}& \textbf{0.912}&\underline{0.967}&\underline{0.984}& \\
			





			

			
			

			\midrule
			\midrule
			
			\rowcolor{blue!5} \multicolumn{13}{c}{Improved Eigen Test Set\cite{uhrig2017sparsity}} \\

			\multicolumn{1}{l}{Watson et al. \cite{watson2021temporal}}&RN18&RN18&K+CS&192$\times$640  &M+Multi-Fr&\underline{0.064}&0.320&3.187&0.104& 0.946&0.990&\underline{0.995}& \\

			\multicolumn{1}{l}{Guizilini et al. \cite{guizilini2022multi}}&Depthformer&RN18&K+CS&192$\times$640  &M+Multi-Fr&\textbf{0.055}&\underline{0.271}&\underline{2.917}&\underline{0.095}& \underline{0.955}&\underline{0.991}&\textbf{0.998}& \\
			

			\midrule		
			
			\multicolumn{1}{l}{Wang et al. \cite{wangfei_hqdec}}&EffV2s&FBv3&K&128$\times$416  &M&0.074&0.403&3.746&0.121& 0.930&0.986&\textbf{0.997}& \\
			
			\multicolumn{1}{l}{Wang et al. \cite{wangfei_hqdec}$^\ddagger$}&EffV2s&FBv3&K&128$\times$416  &M&0.071&0.384&3.632&0.117& 0.935&0.987&\textbf{0.997}& \\

			\multicolumn{1}{l}{\textbf{SmallDepth(Ours)}}&-&FBv3&K&128$\times$416 &M&0.069&0.372&3.685&0.116& 0.935&0.986&\underline{0.996}& \\
			\multicolumn{1}{l}{\textbf{SmallDepth(Ours)$^\ddagger$}}&-&FBv3&K&128$\times$416 &M&\underline{0.066}&0.359&3.617&\underline{0.113}& 0.939&0.987&\underline{0.996}& \\

			\multicolumn{1}{l}{\textbf{HQDecv2(Ours)}}&EffV2s&FBv3&K&128$\times$416 &M&0.067&\underline{0.355}&\underline{3.536}&\underline{0.113}& \underline{0.941}&\underline{0.988}&\textbf{0.997}& \\
			
			\multicolumn{1}{l}{\textbf{HQDecv2(Ours)}$^\ddagger$}&EffV2s&FBv3&K&128$\times$416  &M&\textbf{0.065}&\textbf{0.339}&\textbf{3.440}&\textbf{0.109}& \textbf{0.946}&\textbf{0.989}&\textbf{0.997}& \\

			\midrule

			\multicolumn{1}{l}{Godard et al. \cite{godard2019digging}}&RN18&RN18&K&192$\times$640  &M&0.090&0.545&3.942&0.137& 0.914&0.983&0.995& \\
			
			\multicolumn{1}{l}{Guizilini et al. \cite{guizilini20203d}}&PackNet&PN7*&K+CS&192$\times$640  &M&0.078&0.420&3.485&0.121& 0.931&0.986&0.996& \\
			
			\multicolumn{1}{l}{Zhou et al. \cite{zhou2021self}}&HR18&RN18&K&192$\times$640  &M&0.076&0.414&3.493&0.119& 0.936&0.988&\underline{0.997}& \\
			
			\multicolumn{1}{l}{Zhao et al. \cite{zhao2022monovit}}&MPVit&RN18&K&192$\times$640 &M&0.075&0.389&3.419&0.115& 0.938&0.989&\underline{0.997}& \\
			
			\multicolumn{1}{l}{He et al. \cite{he2022ra}}&HR18&RN18&K&192$\times$640 &M&0.074&0.362&3.345&0.114& 0.940&\underline{0.990}&\underline{0.997}& \\


			\multicolumn{1}{l}{Wang et al. \cite{wangfei_hqdec}}&EffV2s&FBv3&K&192$\times$640  &M&0.065&0.328&3.289&0.107& 0.945&\underline{0.990}&\underline{0.997}& \\
			
			\multicolumn{1}{l}{Wang et al. \cite{wangfei_hqdec}$^\ddagger$}&EffV2s&FBv3&K&192$\times$640  &M&{0.062}&{0.318}&{3.231}&{0.105}& {0.948}&\underline{0.990}&\underline{0.997}& \\

			

			\multicolumn{1}{l}{\textbf{SmallDepth(Ours)}}&-&FBv3&K&192$\times$640 &M&0.067&0.362&3.667&0.114& 0.937&0.986&0.996& \\
			\multicolumn{1}{l}{\textbf{SmallDepth(Ours)$^\ddagger$}}&-&FBv3&K&192$\times$640 &M&0.065&0.350&3.603&0.112& 0.940&0.987&0.996& \\

			\multicolumn{1}{l}{\textbf{HQDecv2(Ours)}}&EffV2s&FBv3&K&192$\times$640  &M&\underline{0.059}&\underline{0.283}&\underline{3.004}&\underline{0.097}& \underline{0.958}&\textbf{0.992}&\textbf{0.998}& \\
			\multicolumn{1}{l}{\textbf{HQDecv2(Ours)}$^\ddagger$}&EffV2s&FBv3&K&192$\times$640 &M&\textbf{0.057}&\textbf{0.274}&\textbf{2.952}&\textbf{0.095}& \textbf{0.961}&\textbf{0.992}&\textbf{0.998}& \\

			\bottomrule %
  	\end{tabular}}
	\caption{ Performance comparison (80 m) on the KITTI dataset. The prediction results were aligned by the median ground-truth LiDAR information. `M'/`S': self-supervised mono/stereo supervision. `Multi-Fr.' indicates that the depth map was predicted by utilizing multiple frames during the inference process. `$^\ddagger$' indicates that the prediction results were aligned by AdaSearch \cite{wangfei_hqdec}. `DE/PE' refers to the backbone of the encoder used in the depth/pose estimation network. The weights of SmallDepth were updated based on $L_{py}$+$L_{apx}$(the pretrained HQDecv2 with $192\times 640$ was used as the large model). } \label{Tab:depth_compared_previous_method_80m} 
	\vspace{-20pt}
\end{table*}
\begin{table*}[htbp]%
	\setlength\tabcolsep{2pt}
	\centering	
	\resizebox{0.75\linewidth}{!}{	
		\begin{tabular}{lccccccccccccccccccc} 
			\toprule %
			
			
			
			{\footnotesize Method}&{\footnotesize DE}&{\footnotesize PE}&{\footnotesize Sup + Multi-Fr?}&{\footnotesize RES}&\footnotesize AbsRel$\downarrow$&\footnotesize SqRel$\downarrow$&\footnotesize RMSE$\downarrow$&\footnotesize RMSElog$\downarrow$& \footnotesize $\delta_1$$\uparrow$ &\footnotesize $ \delta_2$$\uparrow$&\footnotesize $\delta_3$$\uparrow$&	\\ 
			
			\midrule	
			
			\multicolumn{1}{l}{{Guizilini et al.} \cite{guizilini20203d}}&PackNet&PN7*&M&{384$\times$640} &{0.162}&{3.917}&{13.452}&{0.269}& {0.823}&{-}&{-}&\\

			\multicolumn{1}{l}{Han et al. \cite{han2022transdssl}}&Swin&PN7&M&384$\times$640 &0.151&3.591&14.350&0.244&-&-&-&\\
			
			\multicolumn{1}{l}{Guizilini et al. \cite{guizilini2021geometric}}&RN101&RN18&M+Semantic&384$\times$640&0.147&2.922&14.452&-& 0.809&-&-&\\
			
			\multicolumn{1}{l}{Watson et al. \cite{watson2021temporal}}&RN18&RN18&M+Multi-Fr.&-&0.146&3.258&14.098&- &0.822&-&-&\\
			
			\multicolumn{1}{l}{Guizilini et al. \cite{guizilini2022multi}}&Depthformer&RN18&M+Multi-Fr.&-&0.135&2.953&12.477&-& 0.836&-&-&\\

			\multicolumn{1}{l}{Wang et al. \cite{wang2022cbwloss}}&RN50&PN7&M&384$\times$640&
			{0.121}&{1.229}&{6.721}&{0.187}& {0.853}&{0.958}& \textbf{0.985}&\\
			
			\multicolumn{1}{l}{HQDec \cite{wangfei_hqdec}}&EffV2s&FBv3&M&384$\times$640&{0.113}&{1.214}&{6.286}&{0.176}& {0.876} &\underline{0.963}&\textbf{0.985}&\\
			
			\multicolumn{1}{l}{HQDec \cite{wangfei_hqdec}$^\ddagger$}&EffV2s&FBv3&M&384$\times$640&\underline{0.107}&\textbf{1.173}&\textbf{6.109}&\textbf{0.171}& \underline{0.885}&\textbf{0.964}&\textbf{0.985}&\\

			\multicolumn{1}{l}{\textbf{SmallDepth(Ours)}}&-&FBv3&M&384$\times$640&0.118&1.229&6.745&0.188& 0.860&0.957&0.983&\\
			\multicolumn{1}{l}{\textbf{SmallDepth(Ours)$^\ddagger$}}&-&FBv3&M&384$\times$640&0.112&\underline{1.177}&6.547&0.182& 0.867&0.959&\underline{0.984}&\\
			
			\multicolumn{1}{l}{\textbf{HQDecv2(Ours)}}&EffV2s&FBv3&M&384$\times$640&0.108&1.279&6.345&0.176& 0.884&\underline{0.963}&0.983&\\
			\multicolumn{1}{l}{\textbf{HQDecv2(Ours)}$^\ddagger$}&EffV2s&FBv3&M&384$\times$640&\textbf{0.103}&1.246&\underline{6.210}&\underline{0.172}& \textbf{0.894}&\textbf{0.964}&\underline{0.984}&\\

			\bottomrule %
	\end{tabular}}	
	\caption{Monocular depth estimation performance comparison conducted on the DDAD dataset \cite{guizilini20203d}. The weights of SmallDepth were updated based on $L_{cr}$+$L_{apx}$(the pretrained HQDecv2 with $384\times 640$ was used as the large model). }\label{Tab:depth_compared_on_ddad}
	\vspace{-0.4cm}
\end{table*}

\begin{figure*}[htbp]
	\centering	
	\rotatebox{90}{ \textbf{{\scriptsize Ours-A}}  \textbf{{\scriptsize Ours-B}}  \cite{wangfei_hqdec}
		\; \cite{zhao2022monovit}
		\; \cite{guizilini20203d}
		\; \cite{godard2019digging}\quad {\scriptsize GT}}	
	\includegraphics[scale=0.066]{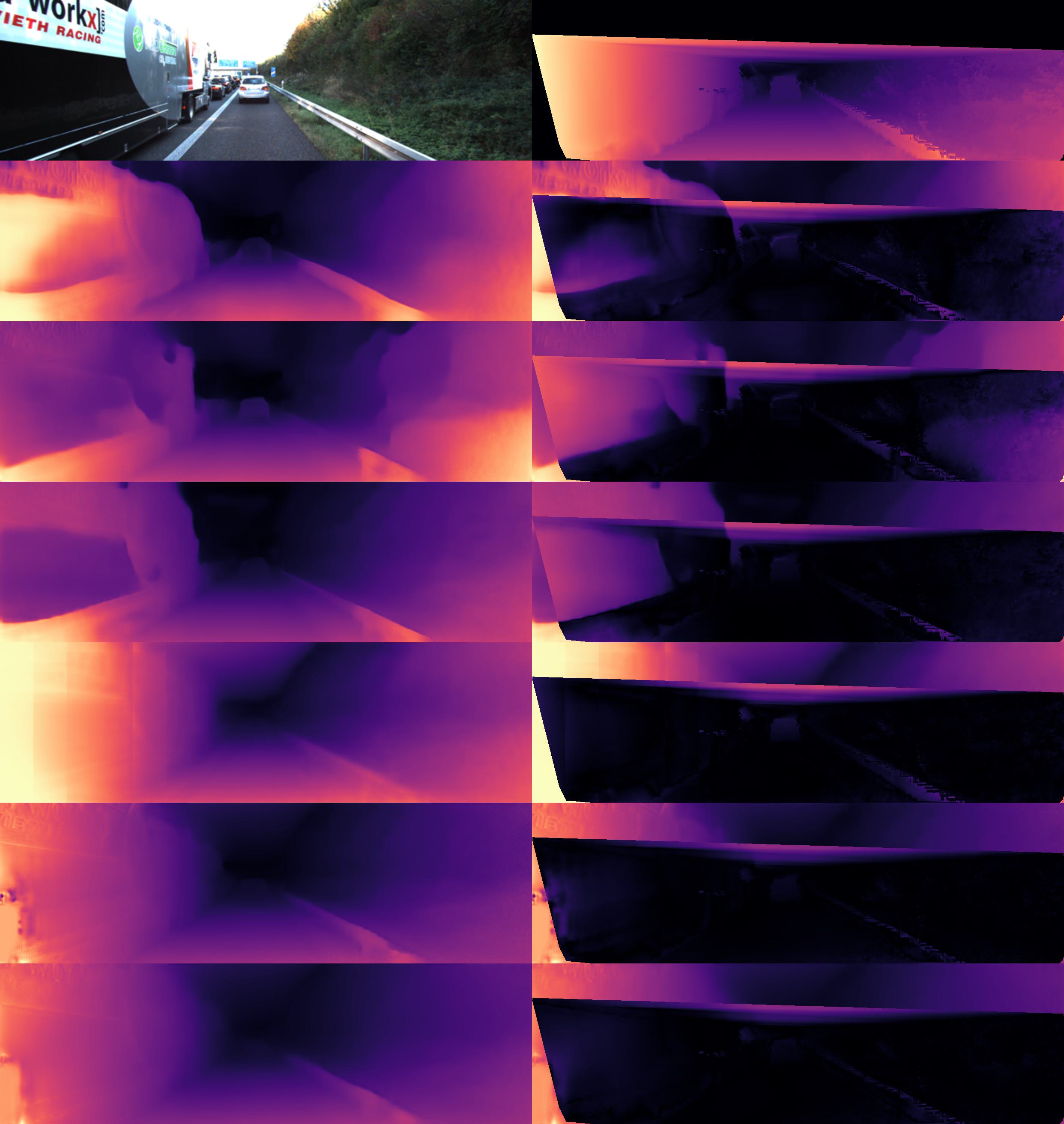}	
	\includegraphics[scale=0.066]{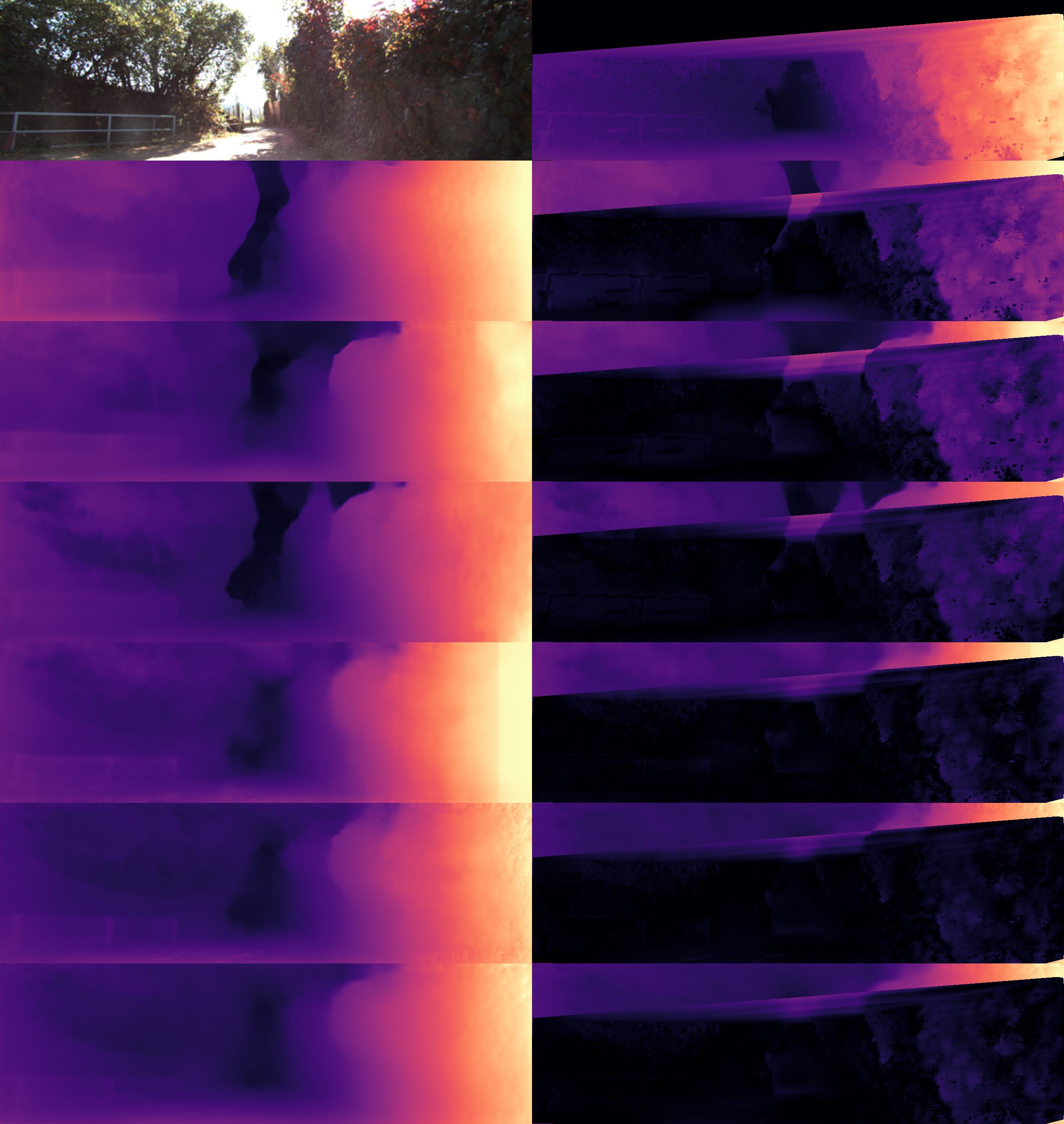}
	\includegraphics[scale=0.066]{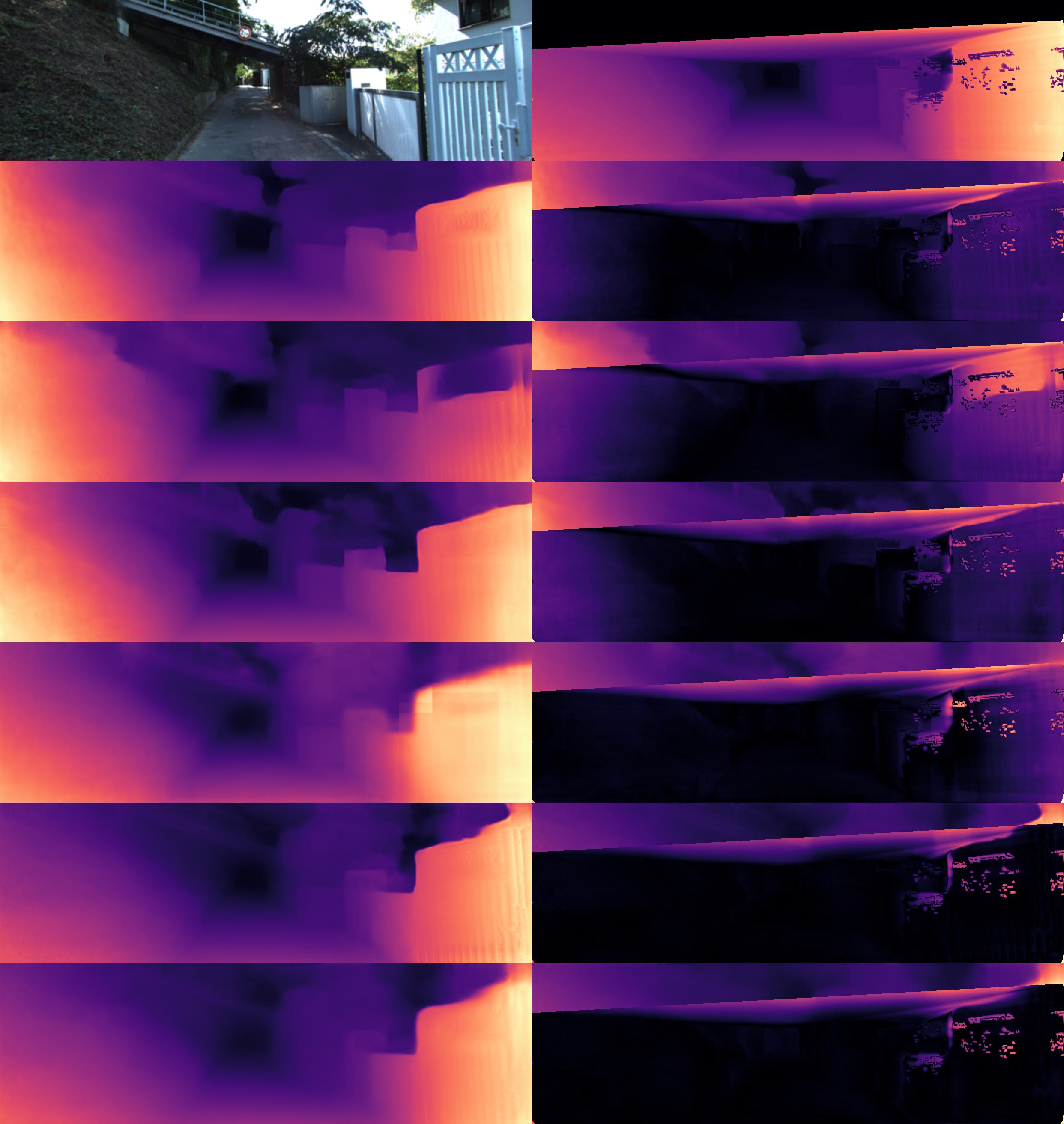}

	\caption{Qualitative comparison ($192\times 640$ ) on KITTI Dataset. Ours-A:SmallDepth, Ours-B: HQDecv2.}
	\label{fig:compare_kitti}
\end{figure*}

Fig. \ref{fig:compare_kitti} displays a qualitative comparison with the previously developed methods \cite{godard2019digging,guizilini20203d,zhao2022monovit,wangfei_hqdec}.  As shown in Fig. \ref{fig:compare_kitti}, in some challenging scenarios(e.g., the truck/tree in the left/remaining sample, the rail in the third column), the previous methods \cite{godard2019digging,guizilini20203d,zhao2022monovit} could not accurately predict the depth of the corresponding region, while the depth predicted by both HQDec \cite{wangfei_hqdec} and the proposed HQDecv2 is close to the groundtruth in these regions, which should benefit from the ability to model local and global dependencies between pixels in parallel. However, since HQDec \cite{wangfei_hqdec} uses non-overlapping sampling strategy to model global dependencies between pixels, resulting in grid artifacts in the predicted depth and tending to infer similar depth from the whole block region with the same size as the subfeature map,  while HQDecv2 effectively eliminate the above phenomenon by using multi-level overlapping sampling to model global information, resulting in a clearer and smoother depth map. Benefiting from the highly accuracy prediction results, the small model also estimated relatively accurate depths in these regions.

\vspace{-10pt}

\subsection{Ablation studies}

\begin{table}[htbp]\footnotesize
	\scriptsize 
	\setlength\tabcolsep{2pt}
	\centering
	\small	
	\resizebox{\linewidth}{!}{
		\begin{tabular}{lccccccccccccccccc} 
			\toprule 
			
			{\footnotesize Scheme}&LR&\scriptsize AbsRel$\downarrow$&\scriptsize SqRel$\downarrow$&\scriptsize RMSE$\downarrow$&\scriptsize RMSE log$\downarrow$&\scriptsize $\delta_1$$\uparrow$&\scriptsize $\delta_2$$\uparrow$&\scriptsize $\delta_3$$\uparrow$& \\ 
			
			\midrule		
			\multicolumn{1}{l}{{\scriptsize RN18}}&{$1e^{-4}$}&0.1203&0.8990&4.9958&0.2019&0.8546&0.9507&0.9783&\\
			
			\multicolumn{1}{l}{{\scriptsize MobileNetV3}}&$1e^{-4}$ &0.1242&0.9167&5.1291&0.2045&0.8451&0.9454&0.9777&\\

			\multicolumn{1}{l}{{\scriptsize SmallDepth(Ours)}}&$1e^{-4}$ &0.1216&0.8990&5.0984&0.2003&0.8517&0.9498&0.9789&\\

			\midrule

			\multicolumn{1}{l}{{\scriptsize RN18}}&$2e^{-4}$&0.1171&0.8667&4.9407&0.1972&0.8614&0.9525&0.9795&\\
			\multicolumn{1}{l}{{\scriptsize MobileNetV3}}&$2e^{-4}$&0.1235&0.9276&5.0207&0.2030&0.8503&0.9483&0.9783&\\

			\multicolumn{1}{l}{{\scriptsize SmallDepth(Ours)}}&$2e^{-4}$&0.1187&0.8708&5.0365&0.1982&0.8568&0.9510&0.9798&\\
			
			\midrule

			\multicolumn{1}{l}{{\scriptsize RN18}}&$4e^{-4}$&0.1169&0.8585&4.9246&0.1967&0.8620&0.9531&0.9798&\\
			\multicolumn{1}{l}{{\scriptsize MobileNetV3}}&$4e^{-4}$&0.1202&0.8889&5.0067&0.1994&0.8560&0.9503&0.9786&\\
			\multicolumn{1}{l}{{\scriptsize SmallDepth(Ours)}}&$4e^{-4}$&0.1157&0.8511&4.9284&0.1952&0.8641&0.9537&0.9799&\\

			\midrule
			
			\multicolumn{1}{l}{{\scriptsize RN18}}&$6e^{-4}$&0.1197&0.8707&4.8949&0.1968&0.8590&0.9526&0.9801&\\
			\multicolumn{1}{l}{{\scriptsize MobileNetV3}}&$6e^{-4}$&0.1264&0.8843&5.0979&0.2020&0.8377&0.9484&0.9805&\\
			\multicolumn{1}{l}{{\scriptsize SmallDepth(Ours)}}&$6e^{-4}$&0.1178&0.8698&4.8766&0.1953&0.8626&0.9546&0.9803&\\

			\midrule
			
			\multicolumn{1}{l}{{\scriptsize RN18}}&$8e^{-4}$&0.1600&1.2062&6.0648&0.2396&0.7647&0.9218&0.9721&\\
			\multicolumn{1}{l}{{\scriptsize MobileNetV3}}&$8e^{-4}$&0.1905&1.4066&6.1680&0.2562&0.7163&0.9079&0.9688&\\
			\multicolumn{1}{l}{{\scriptsize SmallDepth(Ours)}}&$8e^{-4}$&0.1161&0.8544&4.9558&0.1965&0.8615&0.9534&0.9799&\\

			\midrule
			
			\multicolumn{1}{l}{{\scriptsize RN18}}&$10e^{-4}$&0.1647&1.2153&6.3286&0.2439&0.7545&0.9137&0.9705&\\
			\multicolumn{1}{l}{{\scriptsize MobileNetV3}}&$10e^{-4}$&0.2015&1.4863&6.7226&0.2758&0.6646&0.8912&0.9626&\\
			\multicolumn{1}{l}{{\scriptsize SmallDepth(Ours)}}&$10e^{-4}$&0.1168&0.8586&4.9893&0.1978&0.8604&0.9521&0.9795&\\

			\bottomrule 
	\end{tabular}}
	
	\vspace{2pt}
	\resizebox{\linewidth}{!}{ 
	\begin{tabular}{lccccccccccccccccc}		 
		\toprule
		 
		{\scriptsize Scheme}&{\scriptsize FPS}&{\scriptsize GPU-Utili}&N&\makecell{{\scriptsize Param(M)}\\{\scriptsize Enc/Dec}} &\makecell{{\scriptsize GFLOPs}\\{\scriptsize Enc/Dec}}&\makecell{{\scriptsize MAC(M)}\\{\scriptsize Enc/Dec}}\\
		\midrule	
		{\scriptsize RN18}&214&30\%&8&11.18/3.15&1.93/1.55&103.07/59.7\\
		{\scriptsize MobileNetV3}&186&28\%&15&2.82/1.87&0.23/1.09&105.48/50.7\\
		{\scriptsize SmallDepth(Ours)}&557&8\%&6&2.07/0.28&0.25/0.17&39.07/54.5\\
		\bottomrule 
		
	\end{tabular}}
	\caption{Ablation studies on the different architectures. }\label{Tab:ablation_arch_combine_table_different_lr}
	\vspace{-10pt} 
\end{table}

Tab. \ref{Tab:ablation_drop_scheme} shows the effect of different drop strategies on model performance. 
The results show that compared with the schemes without the drop strategy and the schemes with a constant drop rate, the proposed drop strategy, in which the information was randomly dropped with varying drop rate during training is more helpful for improving the generalization ability of the model.

In Tab. \ref{Tab:ablation_arch_combine_table_different_lr}, we investigated the robustness of different models to the learning rate. The proposed SmallDepth can converge well under different learning rates while the DepthNet based on `RN18'/'MobileNetV3' diverges at higher learning rates. This may be because the multi-branch parallel design strategy, adopted by the proposed SmallDepth, reduces the risk of gradient disappearance or gradient explosion. Moreover, from these experimental results, we also obtained a learning rate suitable for SmallDepth.  It can also be seen that SmallDepth not only has lower complexity and faster speed than DepthNet based on `RN18'/'MobileNetV3' but can also  achieve similar or even superior performance.

\begin{table}[htbp]\scriptsize
	\scriptsize 
	\setlength\tabcolsep{2pt} 
	\centering
	\small	
	\resizebox{\linewidth}{!}{
	\begin{tabular}{l|lcccccccccccccccc} 
		\toprule 
		
		&Scheme&$pb_{dsr}$&$pb_{sd}$&AbsRel$\downarrow$&SqRel$\downarrow$&RMSE$\downarrow$&RMSE log$\downarrow$&$\delta_1$$\uparrow$&$\delta_2$$\uparrow$&$\delta_3$$\uparrow$& \\

		\midrule	
		 \multirow{8}{*}{\rotatebox{90}{$LR=1e^{-4}$} }
		&sch1&0.0&0.0&0.1241&0.9335&5.1246&0.2048&0.8484&0.9463&0.9769&\\
		&sch2&0.1&0.0&0.1231&0.9110&5.1291&0.2030&0.8485&0.9469&0.9785&\\
		&sch2&0.5&0.0&0.1221&0.9188&5.0954&0.2014&0.8523&0.9479&0.9782&\\
		&sch3&0.9&0.0&0.1216&0.8990&5.0984&0.2003&0.8517&0.9498&0.9789&\\
		&sch4$^\star$&0.9&0.0&0.1240&0.8894&5.1689&0.2034&0.8446&0.9480&0.9792&\\

		\cline{2-12}
		
		&sch6&0.9&0.1&0.1212&0.8702&5.0570&0.1999&0.8513&0.9491&0.9799&\\
		&sch7&0.9&0.5&0.1210&0.8739&5.0771&0.1993&0.8501&0.9505&0.9800&\\
		&sch8&0.9&0.9&0.1240&0.9053&5.1691&0.2028&0.8453&0.9477&0.9790&\\

		\bottomrule 
	\end{tabular}}
	\caption{Ablation studies on the drop rates $pb_{dsr}$ and $pb_{sd}$ in formulas \eqref{eq_spres_out} and \eqref{eq_complex_spd}. `$\star$' represents the constant drop rate.}\label{Tab:ablation_drop_scheme}
	\vspace{-10pt} 
\end{table}

To shorten the experimental period, the corresponding models in Tab \ref{Tab:ablation_eq_module_hyper_param}, \ref{Tab:ablation_current_flip_low_high_wo_feat_disp}, \ref{Tab:ablation_lpy},  were trained on only the KITTIRAW dataset from scratch with a batch size of 8 and learning rate of $1e^4$ for 50,000 iterations. Although such results do not represent the final results of the model because the model does not converge to saturation, they can reflect the extent to which each module contributes to the performance improvement. It can be seen from the results (see the `SmallDepth' and `SmallDepth+ETM*' schemes in Tab. \ref{Tab:ablation_eq_module_hyper_param}) that the performance of the model will be improved if the filters with other shapes except for the filter with the largest shape are randomly dropped with the appropriate drop rate(e.g. $pb_t=0.1$) during training.  Compared with the `SmallDepth+ETM*' scheme, the  `SmallDepth+ETM' scheme(e.g., $pb_t=0.1$, $r_1=0.1$, $r_2=0.5$), in which dropconv in formula \eqref{eq_drop_conv} was also utilized, achieved better generalization performance.

\begin{table}[htbp]\footnotesize
	\scriptsize 
	\setlength\tabcolsep{2pt}
	\centering
	\small	
	\resizebox{1\linewidth}{!}{
		\begin{tabular}{l|lcccccccccccccccc} 
			\toprule 

			&{\footnotesize Scheme}&Sup&\scriptsize AbsRel$\downarrow$&\scriptsize SqRel$\downarrow$&\scriptsize RMSE$\downarrow$&\scriptsize RMSE log$\downarrow$&\scriptsize $\delta_1$$\downarrow$&\scriptsize $\delta_2$$\downarrow$&\scriptsize $\delta_3$$\downarrow$& \\

			\midrule	
			\multirow{6}{*}{\rotatebox{90}{$LR=4e^{-4}$} }	
			&\multicolumn{1}{l}{{SmallDepth}}&$L_{cr}$&0.1157&0.8511&4.9284&0.1952&0.8641&0.9537&0.9799&\\

			&\multicolumn{1}{l}{{SmallDepth+ETM}}&$L_{cr}$&0.1125&0.8330&4.8683&0.1946&0.8680&0.9543&0.9796&\\
			
			&\multicolumn{1}{l}{{SmallDepth}+ETM$^\#$}&$L_{cr}$&0.1114&0.8322&4.8530&0.1936&0.8714&0.9547&0.9794&\\

			&\multicolumn{1}{l}{RN18+ETM}&$L_{cr}$&0.1106&0.8121&4.7515&0.1915&0.8750&0.9561&0.9801&\\
			&\multicolumn{1}{l}{MobileNetV3+ETM}&$L_{cr}$&0.1121&0.8211&4.8078&0.1941&0.8720&0.9546&0.9797&\\

			&SmallDepth+ETM&$L_{py}$ &0.1071&0.8052&4.6979&0.1878 &0.8809&0.9567&0.9803\\

			\bottomrule 
	\end{tabular}}
	\caption{Ablation studies on ETM.   `ETM$^\#$' represents the weight of ETM initialized by the corresponding weight of the trained SmallDepth. }\label{Tab:ablation_arch_lr4e_4_eq_module}
	\vspace{-20pt} 
\end{table}

To further verify the actual effect of the ETM in Sec. \ref{sec:eq_trans_module} on the model performance,  we utilized the proposed ETM instead of the corresponding filter in different DepthNets (e.g., SmallDepth, DepthNet based on `RN18'/`MobileNetV3' in Tab. \ref{Tab:ablation_arch_lr4e_4_eq_module}). Tab. \ref{Tab:ablation_arch_lr4e_4_eq_module} shows that using the proposed ETM instead of filters helps to improve the inference accuracy of the model without changing the complexity of the model during inference. Compared with `SmallDepth+ETM', the ETM can provide a greater performance boost to DepthNet based on `RN18' and `MobileNetV3' (corresponding to the `RN18+ETM' and `MobileNetV3+ETM' schemes). We hypothesize that this may be because DepthNet based on `RN18' and `MobileNetV3' has more parameters (see Tab. \ref{Tab:ablation_arch_combine_table_different_lr})  than SmallDepth, and ETM can add more capacity to them. Furthermore, if the filter weights of each shape in the ETM were initialized by the weights of the filter with the largest shape, the performance of the algorithm could be further improved (see `SmallDepth+ETM' and `SmallDepth+ETM$^\#$ in Tab. \ref{Tab:ablation_arch_lr4e_4_eq_module}).

\begin{table}[htbp]\footnotesize
	\scriptsize 
	\setlength\tabcolsep{2pt}
	\centering
	\small	
	\resizebox{1\linewidth}{!}{
	\begin{tabular}{l|lcccccccccccccccc} 
		\toprule 
		
		\multicolumn{1}{l}{}&			
		\multicolumn{1}{l}{Scheme}&
		\multicolumn{1}{l}{$pb_{t}$}&
		\multicolumn{1}{l}{$r_1$}&
		\multicolumn{1}{l}{$r_2$}&
		\multicolumn{1}{l}{{\scriptsize AbsRel}}&
		\multicolumn{1}{l}{{\scriptsize SqRel}} &
		\multicolumn{1}{l}{{\scriptsize RMSE}} &
		\multicolumn{1}{l}{{\scriptsize RMSElog}}&
		\multicolumn{1}{l}{\;\;\scriptsize $\delta_1$}\\ 
		
		\midrule
		 
		\multirow{17}{*}{\rotatebox{90}{\makecell{ $LR=1e^{-4}$, BatchSize=8, IterNum=50000 \\ Evaluated on validation set for $128\times 416$.\\ $t\in \{0,..T-1\}$} } }
		
		&SmallDepth &-&-&-& 0.1548	&1.1403&5.8830&0.2273&0.7786\\
		
		&+ETM &0.1&0.1&1 &0.1489&1.0972&5.7242&0.2206&0.7921  \\
		&+ETM&0.05&0.1& 1 & 0.1525&1.1355&5.9215&0.2264&0.7826 \\
		&+ETM&0.1&0.5 &1  &0.1572&1.1854&6.1057&0.2324&0.7707 \\

		\cline{2-10}
		&+ETM&0.1 &0.1 &	0.1 &0.1509& 1.1177&5.8836&0.2250& 0.7855 \\
		&+ETM&0.1&0.1 &0.5  &0.1484&1.0970&5.7794&0.2215&0.7911 \\
		&+ETM&0.1 &0.1 & 1.5  &0.1504& 1.1282&5.8832&0.2244&0.7861 \\
		&+ETM&0.1&0.1  &2  &0.1486&1.1049&5.8234&0.2223&0.7898 \\
		&+ETM&0.1 &0.1  & 0.8  & 0.1491&1.1017&5.7558&0.2212&0.7908\\

		\cline{2-10}
	
		&+ETM$^\divideontimes$&0.0&-&-&0.1542&1.1494&5.9776&0.2286&0.7777 \\
		&+ETM$^\divideontimes$&0.05&-&-&0.1535&1.1433&5.9186&0.2275&0.7807 \\
		&+ETM$^\divideontimes$&0.1&-&-&0.1519&1.1287&5.9584&0.2263&0.7807 \\
		&+ETM$^\divideontimes$&0.2&-&-&0.1576&1.1680&6.0744&0.2326&0.7700 \\
		&+ETM$^\divideontimes$&0.4&-&-&0.1665&1.2649&6.4145&0.2440&0.7484 \\
		&+ETM$^\divideontimes$&0.6&-&-&0.1582&1.1885&6.1931&0.2346&0.7656 \\
		&+ETM$^\divideontimes$&0.8&-&-&0.1555&1.1530&6.0296&0.2307&0.7746 \\
		&+ETM$^\divideontimes$&0.9&-&-&0.1586&0.1713&6.0571&0.2328&0.7687 \\

		\bottomrule 
	\end{tabular}}
	\caption{Ablation studies on the ETM.  `ETM$^\divideontimes$' indicates that the dropconv in formula \eqref{eq_drop_conv} was not used in the ETM. }\label{Tab:ablation_eq_module_hyper_param} 
	\vspace{-10pt} 
\end{table}

\begin{table}[htbp]\footnotesize
	\scriptsize 
	\setlength\tabcolsep{2pt}
	\centering
	\small	
	\resizebox{0.95\linewidth}{!}{
		\begin{tabular}{lccccccccccccccccc} 
			\toprule 
			\multicolumn{1}{l}{{\scriptsize Scheme}}&
			\multicolumn{1}{l}{{\scriptsize cr}}&
			\multicolumn{1}{l}{{\scriptsize flip}}&
			\multicolumn{1}{l}{{\scriptsize lr}}&
			\multicolumn{1}{l}{{\scriptsize hr}}&
			\multicolumn{1}{l}{{\scriptsize color}}&
			\multicolumn{1}{l}{{\scriptsize use\_cr}}&	
			
			\multicolumn{1}{l}{{\scriptsize AbsRel}} & 
			\multicolumn{1}{l}{{\scriptsize SqRel}} &
			\multicolumn{1}{l}{{\scriptsize RMSE}} &
			\multicolumn{1}{l}{{\scriptsize RMSElog}}&
			\multicolumn{1}{l}{\;\;\scriptsize $\delta_1$}\\

			\midrule

			$\boldsymbol{a_1}$&\usym{1F5F8}& & & & & &0.1548&1.1403&5.8830&0.2273&0.7786&\\
			$a_2$&\usym{1F5F8}&\usym{1F5F8}& & & & &0.1458&1.0688&5.5642&0.2156&0.8002  	\\
			$a_3$&\usym{1F5F8}& &\usym{1F5F8}& & & &0.1510&1.1065&5.7296&0.2226&0.7867  	\\
			$a_4$&\usym{1F5F8}& &\usym{1F5F8}& & &\usym{1F5F8}&0.1547&1.1396&5.7315&0.2253&0.7829 	\\
			$a_5$&\usym{1F5F8}& & &\usym{1F5F8}& & &0.1485&1.0814&5.7523&0.2196&0.7935 	\\
			$a_6$&\usym{1F5F8}& & &\usym{1F5F8}& &\usym{1F5F8}&0.1553&1.1464&5.9745&0.2292&0.7760&\\
			
			$a_7$&\usym{1F5F8}& & & &\usym{1F5F8} & &0.1520&1.1297&5.7769&0.2235&0.7859 \\
			
			\bottomrule 
	\end{tabular}}
	\caption{{\small Ablation studies on different combinations of $L_{k}$.}
	}\label{Tab:ablation_current_flip_low_high_wo_feat_disp}
	\vspace{-15pt} 
\end{table}

\begin{table*}[htbp]\footnotesize
	\scriptsize 
	\setlength\tabcolsep{2pt}
	\centering
	\small
	\resizebox{\linewidth}{!}{	
		\begin{tabular}{lccccccccccccccccccccccccccccc} 
			\toprule

			Scheme&$\alpha_{cr}$&$\alpha_{lr}$&$\alpha_{hr}$&$\alpha_{flip}$&$\alpha_{color}$&
			$\beta_{lr}$&$\beta_{hr}$&$\beta_{flip}$&$\beta_{color}$&
			$\gamma_{lr}$&$\gamma_{hr}$&$\gamma_{flip}$&$\gamma_{color}$&
			AbsRel$\downarrow$&SqRel$\downarrow$&RMSE$\downarrow$&RMSE log$\downarrow$&$\delta_1$$\uparrow$&$\delta_2$$\uparrow$&$\delta_3$$\uparrow$& \\ 
			
			\midrule

			w/o $L_{py}$ &1.0&0.0 &0.0&0.0&0.0&0.0&0.0&0.0&0.0&0.0&0.0&0.0&0.0&0.1157&0.8511&4.9284&0.1952 &0.8641&0.9537&0.9799\\
			
			lr&1.0&0.5 &0.0&0.0&0.0&1.0&0.0&0.0&0.0&0.5&0.0&0.0&0.0&0.1130&0.8091&4.8730&0.1934 &0.8684&0.9553&0.9804\\
			
			hr&1.0&0.0 &0.5&0.0&0.0&0.0&0.5&0.0&0.0&0.0&0.1&0.0&0.0&0.1109&0.8300&4.8471&0.1922 &0.8739&0.9565&0.9804\\

			flip* &1.0&0.0 &0.0&1.0&0.0&0.0&0.0&0.1&0.0&0.0&0.0&0.1&0.0&0.1125&0.8518&4.8630&0.1935&0.8720&0.9552&0.9798\\

			flip &1.0&0.0 &0.0&1.0&0.0&0.0&0.0&0.1&0.0&0.0&0.0&0.5&0.0&0.1117&0.8479&4.8682&0.1925&0.8735&0.9560&0.9800\\

			color*  &1.0&0.0&0.0&0.0&1.0&0.0&0.0&0.0&0.1&0.0&0.0&0.0&0.1&0.1118&0.8458&4.9313&0.1954&0.8698&0.9543&0.9789\\
		
			color  &1.0&0.0&0.0&0.0&1.0&0.0&0.0&0.0&0.1&0.0&0.0&0.0&0.1&0.1129&0.8313&4.9993&0.1963 &0.8665&0.9534&0.9792\\

			\bottomrule 
	\end{tabular}}
	\caption{Ablation studies on $L_{py}$. `*' indicates that the L1-norm was used to calculate the feature/disparity errors.  `w/o $L_{py}$' represents the loss function used in \cite{wang2022cbwloss,wangfei_hqdec} was used as the supervisory signal.}\label{Tab:ablation_pyr_loss}
	\vspace{-15pt} 
\end{table*}

\begin{table}[htbp]\footnotesize
	\setlength\tabcolsep{2pt}
	\centering
	\small

	\resizebox{0.95\linewidth}{!}{
		\begin{tabular}{l|lcccccccccccccccc} 
			\toprule 
			\multicolumn{1}{l}{}&			
			\multicolumn{1}{l}{Scheme}&
			\multicolumn{1}{l}{$\alpha_{cr}$}&	
			\multicolumn{1}{l}{$\alpha_{i}$}&	
			\multicolumn{1}{l}{$\beta_{i}$}&	
			\multicolumn{1}{l}{$\gamma_{i}$}&			
			\multicolumn{1}{l}{{\scriptsize AbsRel}}&
			\multicolumn{1}{l}{{\scriptsize SqRel}} &
			\multicolumn{1}{l}{{\scriptsize RMSE}} &
			\multicolumn{1}{l}{{\scriptsize RMSElog}}&
			\multicolumn{1}{l}{\;\;\scriptsize $\delta_1$}\\ 
			
			\midrule
			\multirow{10}{*}{\rotatebox{90}{$i=lr$ }}
			
			&$\boldsymbol{a_3}$& 0.5&0.5&0.0&0.0& 0.1510	&1.1065&5.7296&0.2226&0.7867\\
			&$\boldsymbol{\hat {a}_3}$& 1.0&1.0&0.0&0.0& 0.1526&1.1170&5.8800&0.2263&0.7808\\
			&$\boldsymbol{\hat {a}_{3\_1}}$&1.0&0.1&0.0&0.0 &0.1526&1.1151&5.8179&0.2251&0.7805 \\
			&$\boldsymbol{\hat {a}_{3\_2}}$&1.0&0.5&0.0&0.0 &0.1515&1.1037&5.8290&0.2247&0.7832 \\
			
			\cline{2-11}
			
			&$\hat {a}_{3\_2\_1}$&1.0&0.5&0.1&0.0 &0.1508&1.1045&5.8444&0.2248&0.7841 \\
			&$\hat {a}_{3\_2\_2}$&1.0&0.5&0.5&0.0 &0.1509&1.0963&5.7298&0.2222&0.7866 \\
			&$\hat {a}_{3\_2\_3}$&1.0&0.5&1.0&0.0 &0.1508&1.1084&5.7134&0.2218&0.7896 \\
			
			\cline{2-11}
			
			&$\hat {a}_{3\_2\_4}$&1.0&0.5&0.0&0.1 &0.1517&1.1191&5.8398&0.2251&0.7831 \\
			&$\hat {a}_{3\_2\_5}$&1.0&0.5&0.0&0.5 &0.1491&1.0808&5.7493&0.2215&0.7893 \\
			&$\hat {a}_{3\_2\_6}$&1.0&0.5&0.0&1.0 &0.1506&1.1013&5.6558&0.2210&0.7910 \\

			\bottomrule 
			\bottomrule

			\multirow{10}{*}{\rotatebox{90}{$i=hr$ }}
			&$\boldsymbol{a_5}$& 0.5&0.5&0.0&0.0& 0.1485	&1.0814&5.7523&0.2196&0.7935\\
			
			&$\boldsymbol{\hat {a}_5}$ &1.0&1.0&0.0&0.0&0.1500&1.0991&5.7043&0.2209&0.7923\\
			&$\boldsymbol{\hat {a}_{5\_1}}$&1.0&0.1 &0.0&0.0&0.1538&1.1336&5.8451&0.2259&0.7818 \\
			&$\boldsymbol{\hat {a}_{5\_2}}$&1.0&0.5&0.0&0.0&0.1484&1.0817&5.7614&0.2204&0.7924\\   %
			\cline{2-11}

			&$\hat {a}_{5\_2\_1}$&1.0&0.5&0.1&0.0 &0.1485&1.0896&5.7546&0.2200&0.7931 \\
			& $\hat {a}_{5\_2\_2}$&1.0&0.5&0.5&0.0 & 0.1480&1.0809&5.6468&0.2173&0.7967 \\
			&$\hat {a}_{5\_2\_3}$&1.0&0.5&1.0&0.0 & 0.1488&1.0911&5.7132&0.2195&0.7942 \\
			\cline{2-11}
			
			&$\hat {a}_{5\_2\_4}$&1.0&0.5&0.0&0.1 & 0.1470&1.0740&5.6616&0.2174&0.7972 \\
			&$\hat {a}_{5\_2\_5}$&1.0&0.5&0.0&0.5 & 0.1496&1.0924&5.7920&0.2220&0.7884 \\
			&$\hat {a}_{5\_2\_6}$&1.0&0.5&0.0&1.0 & 0.1519&1.1242&5.9008&0.2256&0.7845 \\
			
			\bottomrule 
			\bottomrule

			\multirow{10}{*}{\rotatebox{90}{$i=flip$ }}
			&$\boldsymbol{a_2}$& 0.5 &0.5&0.0&0.0& 0.1458 &1.0688&5.5642&0.2156&0.8002\\
			&$\boldsymbol{\hat{a}_2}$&1.0&1.0 &0.0&0.0&0.1445&1.0558&5.5148&0.2143&0.8037 \\
			&$\boldsymbol{\hat{a}_{2\_1}}$&1.0&0.5 &0.0&0.0&0.1463&1.0769&5.6634&0.2175&0.7974  \\
			&$\boldsymbol{\hat{a}_{2\_2}}$&1.0&0.1 &0.0&0.0&0.1548&1.1502&5.9854&0.2289&0.7748 \\
			
			\cline{2-11}  
			&$\hat{a}_{2\_0\_1}$&1.0&1.0&0.1&0.0&0.1433&1.0406&5.5965&0.2145&0.8029 \\
			&$\hat{a}_{2\_0\_2}$&1.0&1.0&0.5&0.0&0.1439&1.0386&5.5039&0.2128&0.8041 \\
			&$\hat{a}_{2\_0\_3}$&1.0&1.0&1.0&0.0&0.1436&1.0268&5.5663&0.2143&0.8021 \\
			
			\cline{2-11}
			&$\hat{a}_{2\_0\_4}$&1.0&1.0&0.0&0.1&0.1441&1.0355&5.4914&0.2126&0.8049 \\
			&$\hat{a}_{2\_0\_5}$&1.0&1.0&0.0&0.5&0.1425&1.0268&5.4610&0.2115&0.8072 \\
			&$\hat{a}_{2\_0\_6}$&1.0&1.0&0.0&1.0&0.1442&1.0437&5.4868&0.2129&0.8046 \\
			
			\bottomrule 
			\bottomrule
			
			\multirow{10}{*}{\rotatebox{90}{$i=color$ }}
			&$\boldsymbol{a_7}$& 0.5 &0.5&0.0&0.0& 0.1520&1.1297&5.7769&0.2235&0.7859\\
			
			&$\boldsymbol{\hat{a}_7}$&1.0&0.5&0.0&0.0 &0.1522&1.1246&5.7585&0.2235&0.7759 \\
			&$\boldsymbol{\hat{a}_{7\_1}}$&1.0&1.0&0.0&0.0 &0.1496&1.1038&5.6918&0.2203&0.7908\\
			&$\boldsymbol{\hat{a}_{7\_2}}$&1.0&0.1 &0.0&0.0&0.1573&1.1640&5.8393&0.2282&0.7759\\
			
			\cline{2-11}
			
			&$\hat{a}_{7\_0\_1}$&1.0&1.0&0.1&0.0 &0.1487&1.1001&5.6336&0.2184&0.7949\\
			&$\hat{a}_{7\_0\_2}$&1.0&1.0&0.5&0.0 &0.1563&1.1536&5.8977&0.2290&0.7749\\
			&$\hat{a}_{7\_0\_3}$&1.0&1.0&1.0&0.0 &0.1643&1.2350&6.0935&0.2375&0.7594 \\
			\cline{2-11}
			
			&$\hat{a}_{7\_0\_4}$&1.0&1.0&0.0&0.1 &0.1480&1.0917&5.6682&0.2193&0.7945\\
			&$\hat{a}_{7\_0\_5}$&1.0&1.0&0.0&0.5 &0.1506&1.1173&5.7328&0.2219&0.7888\\
			&$\hat{a}_{7\_0\_6}$&1.0&1.0&0.0&1.0 &0.1532&1.1374&5.7703&0.2242&0.7847 \\
			\bottomrule
			
	\end{tabular}}
	\caption{Ablation studies on $L_{py}$. }\label{Tab:ablation_lpy}
	\vspace{-10pt} 
\end{table}

To better understand the contribution of each component of the proposed pyramid loss in Sec. \ref{sec:pyr_loss} ---  $L_{k}$, $\phi_k(d_{cr},\hat {d}_k)$, $\phi_k(f_{cr},\hat {f}_k)$  $k \in \{cr, lr, hr, flip, color \}$ --- to the overall performance, we performed ablation studies, as shown in Tab. \ref{Tab:ablation_current_flip_low_high_wo_feat_disp}, \ref{Tab:ablation_lpy}.  The results in Tab. \ref{Tab:ablation_current_flip_low_high_wo_feat_disp} show that using the loss calculated from different input samples at the same time to guide the updating of weights is helpful for obtaining better depth predictions. We suspect that this performance gain may be due to the following reasons: compared with the guidance signal obtained from $L_{cr}$ based on the input sample $I_{cr}$ at the current resolution (e.g., $128\times 416$), (1) the signal obtained from $L_{lr}$ based on $I_{cr}$ (e.g., $96\times 320$) can utilize a longer range of context information to guide the updating of weights. (2) the signal obtained from $L_{hr}$ based on $I_{hr}$ (e.g., $160\times 512$) is able to pay more attention to local detail differences and use them to guide the updating of weights. (3) Although people can easily recognize the source image and the flipped image as the same image, DepthNet tends to predict them as different images. The scheme that calculates the mean of $L_{cr}$ and $L_{flip}$ can improve the robustness of DepthNet to the above phenomenon. (4) The signal obtained by calculating the mean of $L_{cr}$ based on  $I_{cr}$ and $L_{color}$ based on $I_{color}$ can reduce noise interference caused by changes in lighting, resulting in improving the robustness of DepthNet to illumination changes. Furthermore, it can be seen from Tab. \ref{Tab:ablation_current_flip_low_high_wo_feat_disp} that the results obtained by not scaling the output feature map and disparity map to the corresponding resolution obtained on `$I_{cr}$' sample (corresponding to scheme `$a_3$, $a_5$') are superior to those obtained by scaling (corresponding to scheme `$a_4$, $a_6$'). This may be because measuring multiple resolution errors simultaneously improves the robustness of DepthNet to scale. To reduce the footprint of video memory, instead of directly summing multiple losses (corresponding to the scheme without the symbol `\^ \;' in Tab. \ref{Tab:ablation_lpy}, we directly sum the gradient calculated by each loss so that the intermediate variables occupying the memory can be released immediately after calculating the gradient, thus reducing the memory usage. These two schemes for updating weights can achieve similar results(e.g., schemes $a_3$ and $\hat{a}_{3\_2}$, schemes $a_5$ and $\hat{a}_{5\_2}$, schemes $a_2$ and $\hat{a}_{2}$, schemes $a_7$ and $\hat {a}_{7\_1}$)  if the appropriate combination coefficients are used. Furthermore, regardless of the scheme of updating weights, combining $L_i, i\in \{lr, hr, color, flip\}$ and $L_{cr}$ can improve the depth prediction performance.

The results on $\beta_{i}$, $\gamma_{i}$, $i\in \{lr, hr, flip, color \}$ in Tab. \ref{Tab:ablation_lpy} show that it is helpful to improve the depth prediction performance by enforcing the consistency between different feature maps or by enforcing consistency between different disparities, obtained from the current sample $I_{cr}$ and the corresponding transformed sample $I_{i}$, respectively. These experimental results also prove that the appropriate combination is more helpful for improving the depth prediction performance.  This may be because while gradient summation can exploit multiple guidance signals simultaneously, it lacks explicit constraints to ensure consistency between the output obtained based on the current input sample $I_{cr}$ and the output obtained based on the transformed sample $I_{i}$, and adding consistency constraints can further limit the solution space.

\begin{table}[htbp]\footnotesize
	\scriptsize 
	\setlength\tabcolsep{2pt}
	\centering
	\small	
	\resizebox{\linewidth}{!}{
		\begin{tabular}{lccccccccccccccccccccccccccccc} 
			\toprule

			Sup&$\lambda_{enc}$&$\lambda_{dec}$&$\lambda_{disp}$&Mask
			&AbsRel&SqRel&RMSE&RMSE log&$\delta_1$&$\delta_2$&$\delta_3$& \\

			\midrule

			$L_{cr}$ &-&-&-&- &0.1114&0.8322&4.8530&0.1936 &0.8714&0.9547&0.9794\\

			$L_{apx}$ &0.0&0.0&1.0&  &0.1025&0.7386&4.9169&0.1858 &0.8748&0.9573&0.9818\\

			$L_{apx}$ &0.01&0.01&1.0& &0.1023&0.7096&4.8442&0.1844 &0.8790&0.9583&0.9823\\

			$L_{apx}$ &0.0&0.0&1.0&\usym{1F5F8} &0.1018&0.7082&4.8067&0.1837 &0.8774&0.9585&0.9822\\

			$L_{apx}$+$L_{cr}$ &0.01&0.01&1.0&\usym{1F5F8} &0.1014&0.7086&4.7161&0.1825 &0.8805&0.9592&0.9824\\

			\bottomrule 
	\end{tabular} }
	\caption{Ablation studies on $L_{apx}$ in Sec. \ref{sec:func_approx_loss_method} for $128\times 416$. The results predicted by HQDecv2 from the $128\times 416$ resolution input sequence were used in $L_{apx}$. `Sup' indicates that the corresponding loss function was used to guide the updating weight of SmallDepth with ETM. }\label{Tab:ablation_func_approx_loss}  
	\vspace{-10pt} 
\end{table}

\begin{figure}[!htbp]
	\resizebox{\linewidth}{!}{
		\rotatebox{90}{{\scriptsize \; \; R4  \; \; \; R3  \quad   R2 \qquad R1}}\hspace{-5pt}
		\subfigure[KITTI Dataset(128x416). `R4': attention maps. `R3': error maps. `R2': disparity maps. `R1':RGB, GT.]{	
			\includegraphics[scale=0.06]{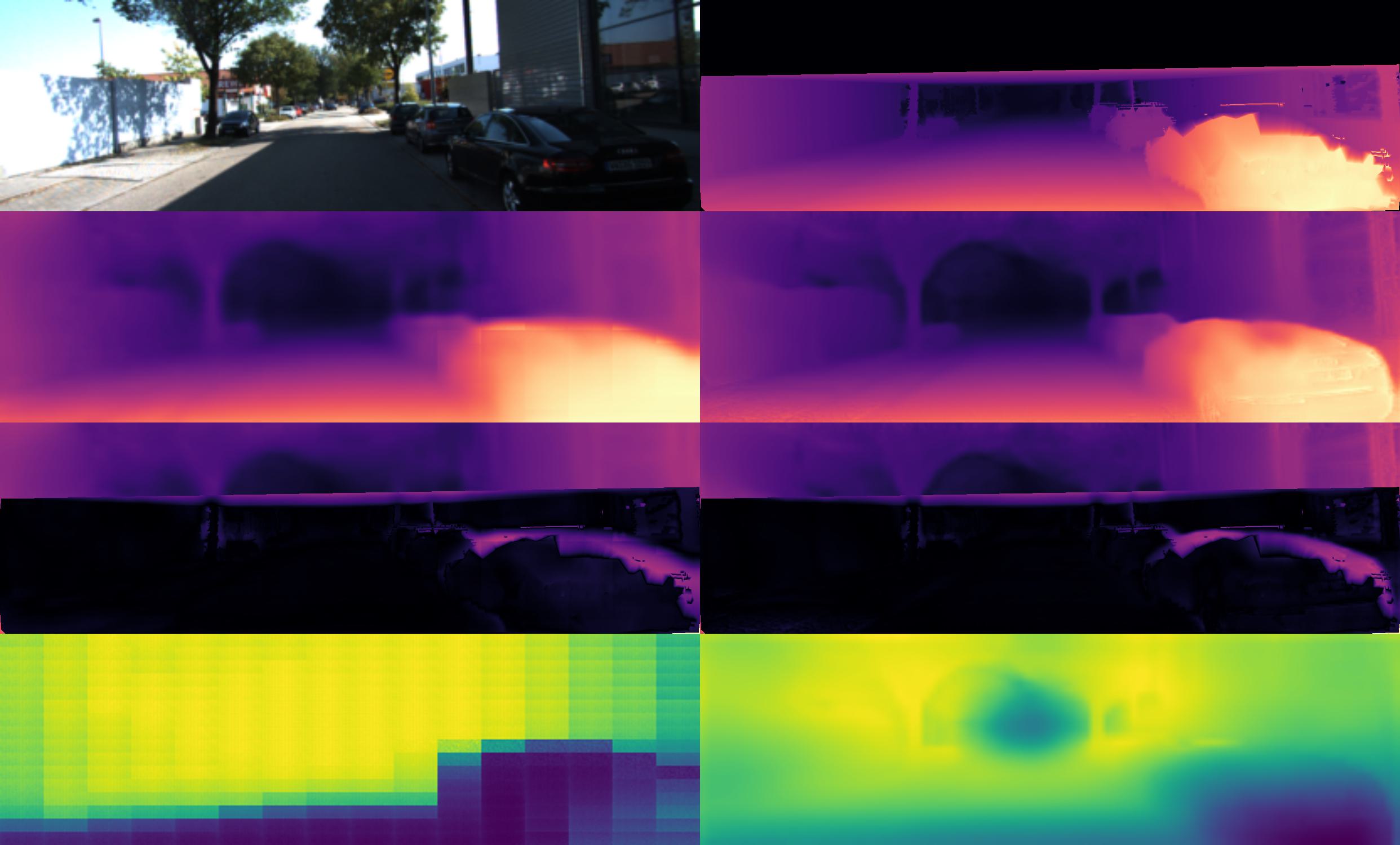}
			\includegraphics[scale=0.06]{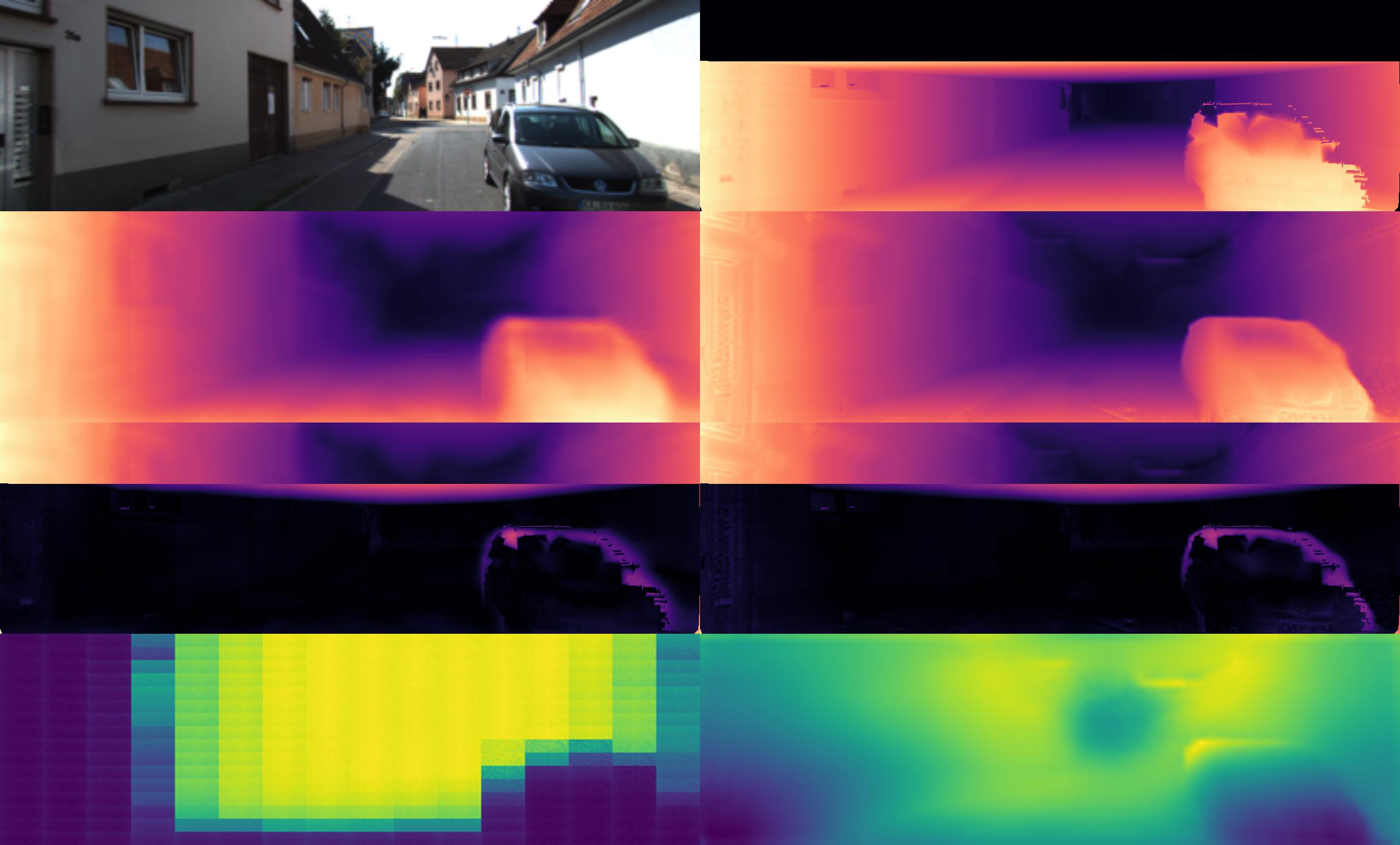}
	}}
	\\
	\vspace{-5pt}
	\resizebox{\linewidth}{!}{
		\subfigure[DDAD Dataset (384x640).]{	
			\includegraphics[scale=0.025]{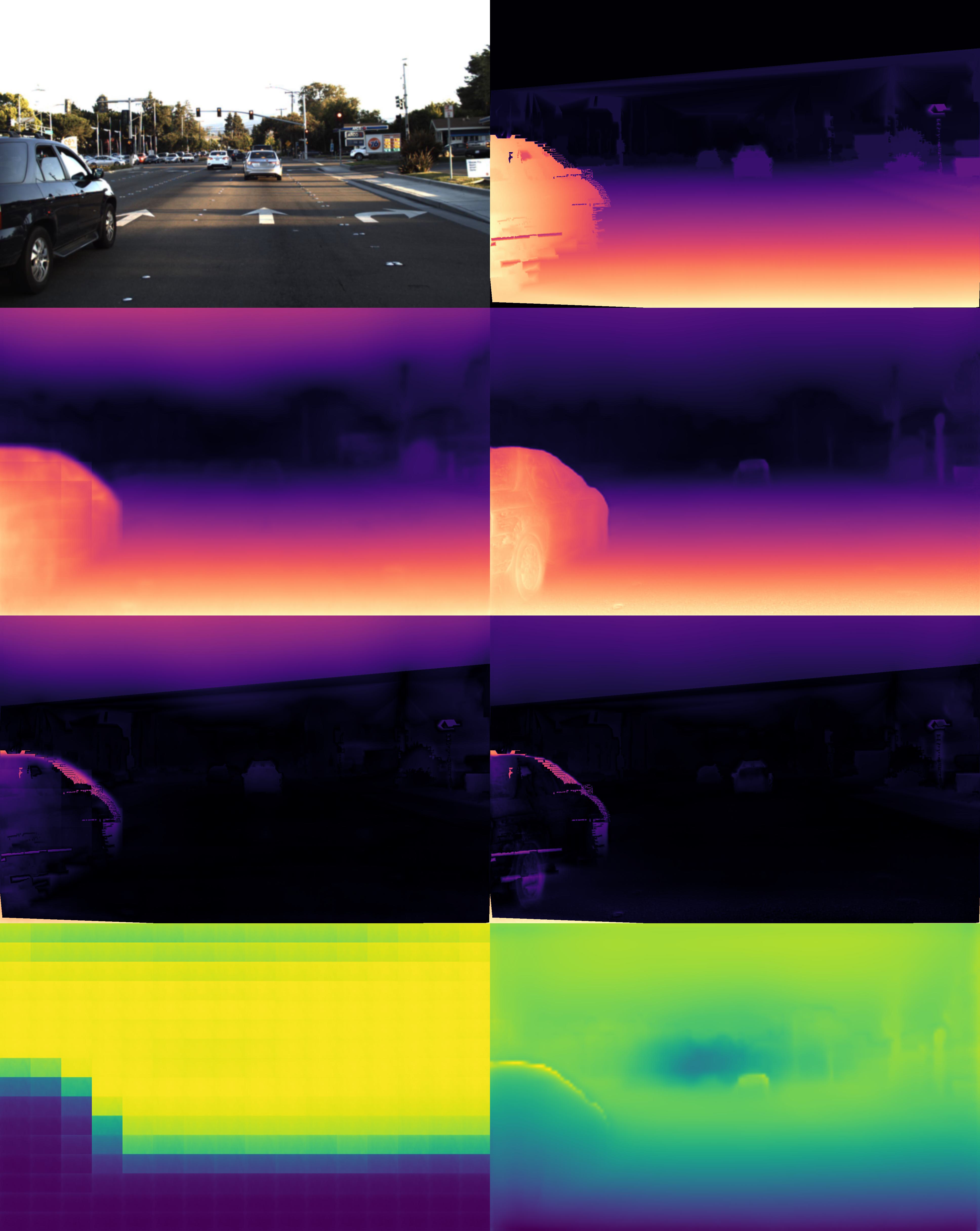}
			\includegraphics[scale=0.025]{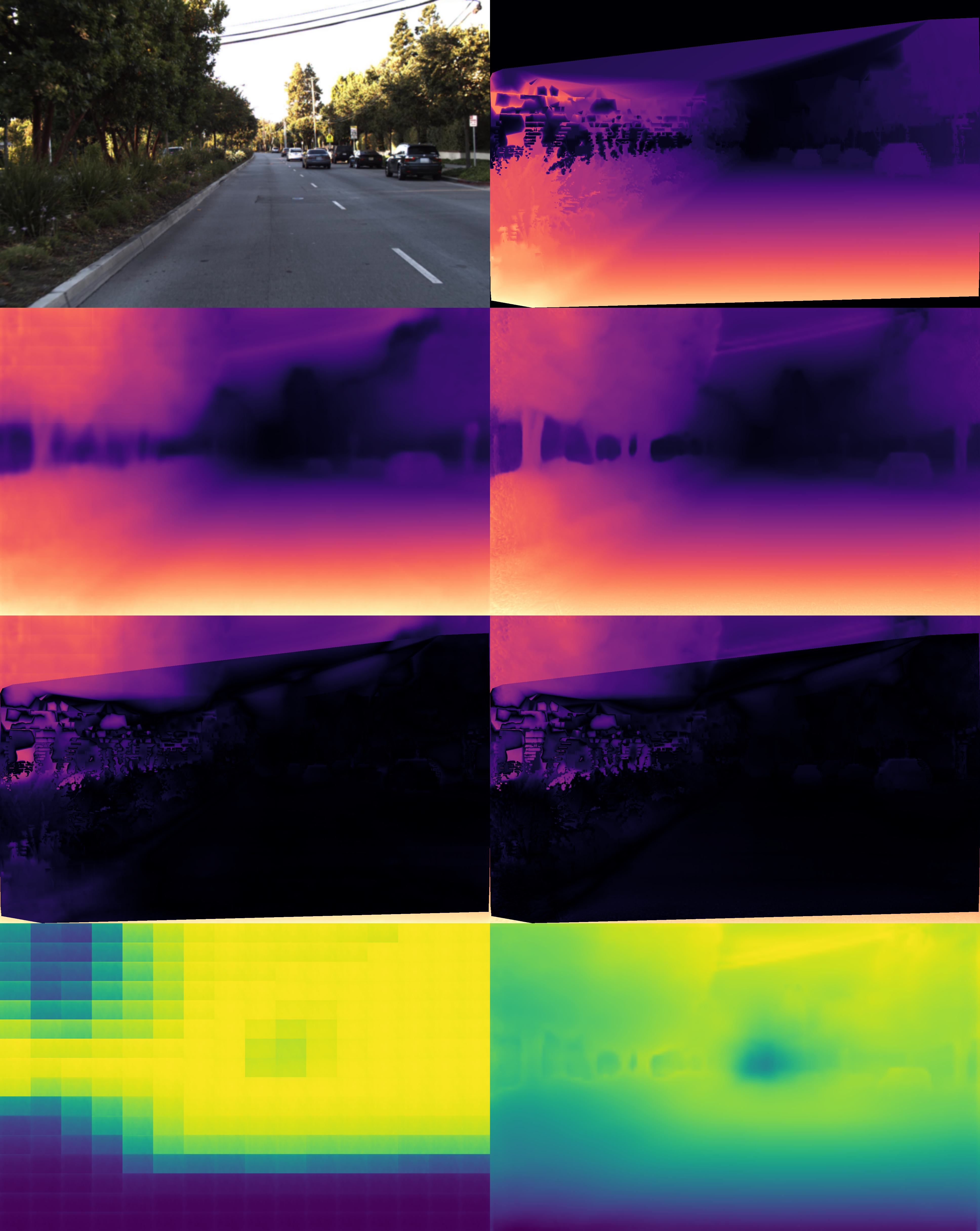}
			\includegraphics[scale=0.025]{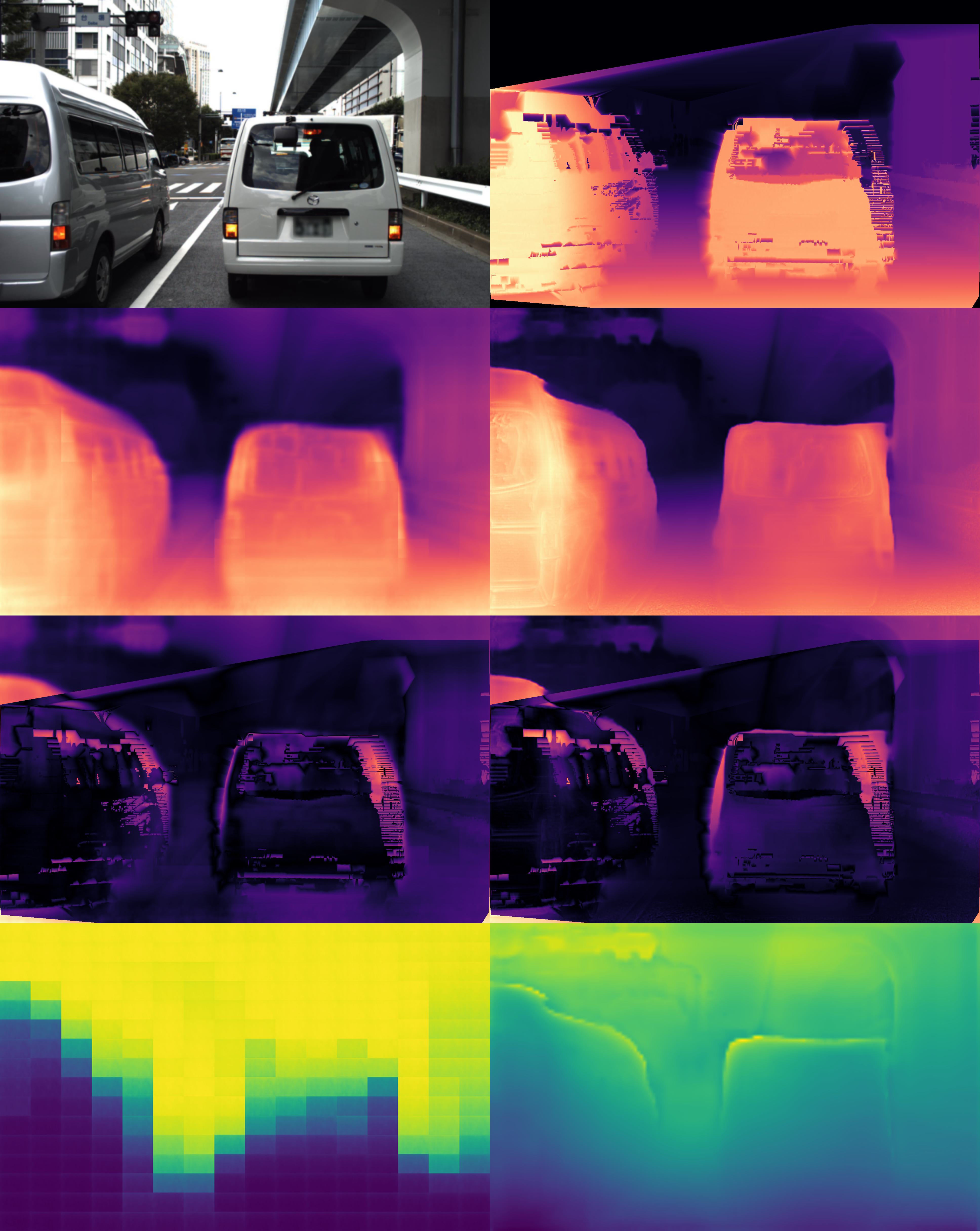}
	}}
	
	\caption{Qualitative comparison between HQDec (left) and HQDecv2 (right). } \label{Fig:hqdec_hqdecv2} 
\end{figure}


\begin{figure*}[!htbp]
	\rotatebox{90}{{\scriptsize \; R3a \quad \quad \; R2 \qquad R1}}\hspace{-5pt}
	\subfigure[The influence of illumination change.  `R3a': scheme `color*'. ]{\label{Fig:lpy_illumination}				
		\includegraphics[scale=0.2]{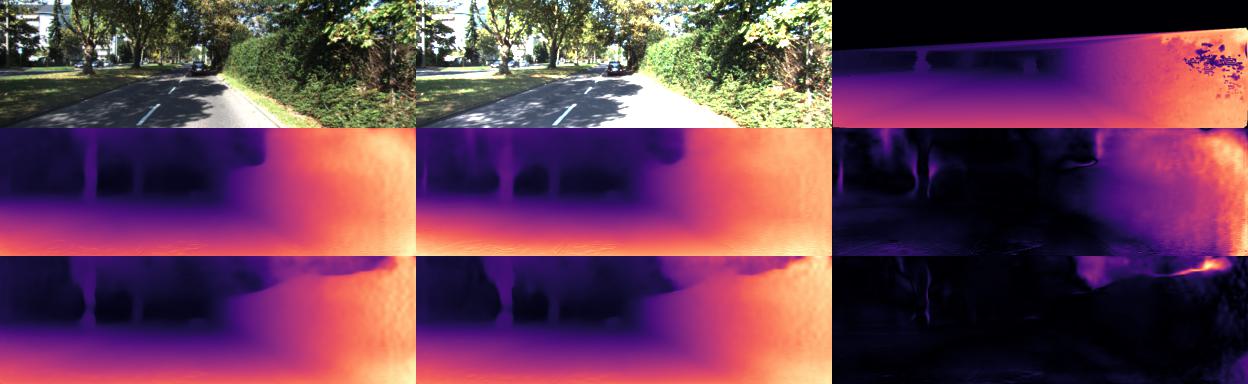}}
	\vspace{-5pt} \;
	\rotatebox{90}{{\scriptsize \; R3b \quad \quad \; R2 \qquad R1}}\hspace{-5pt}
	\subfigure[The influence of flip transformation. `R3b': scheme `flip'.]{\label{Fig:lpy_flip}		
		\includegraphics[scale=0.2]{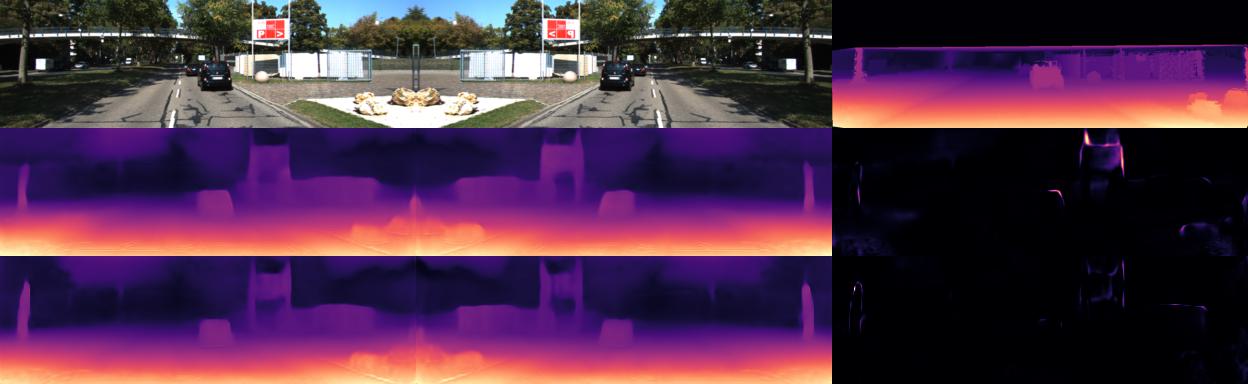}} 
	\\
	\rotatebox{90}{{\scriptsize \; R3c \quad \quad \; R2 \qquad R1}}\hspace{-5pt}
	\subfigure[The influence of low-resolution sample. `R3c': scheme `lr'.]{\label{Fig:lpy_low}	
		\includegraphics[scale=0.2]{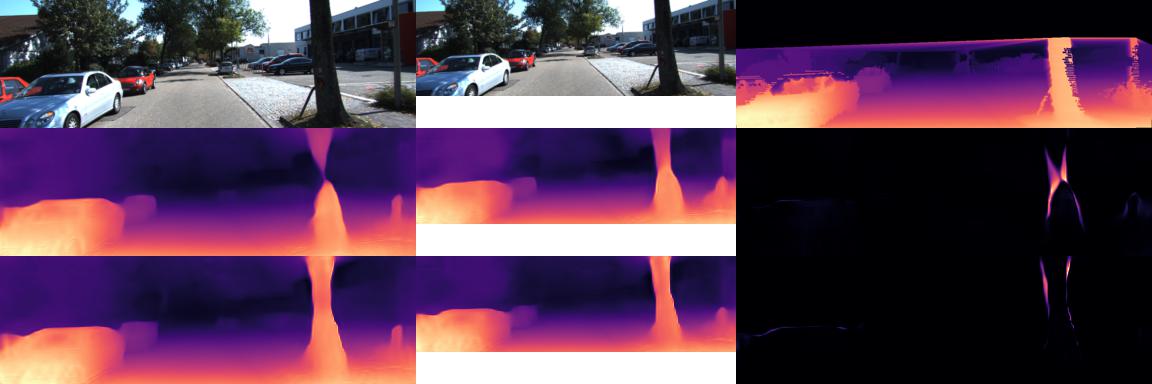}}
	\vspace{-5pt}
	\quad \quad 
	\rotatebox{90}{{\scriptsize \; R3d \quad \quad \; R2 \qquad R1}}\hspace{-5pt}
	\subfigure[The influence of high-resolution sample. `R3d': scheme `hr'.]{\label{Fig:lpy_high}	
		\includegraphics[scale=0.17]{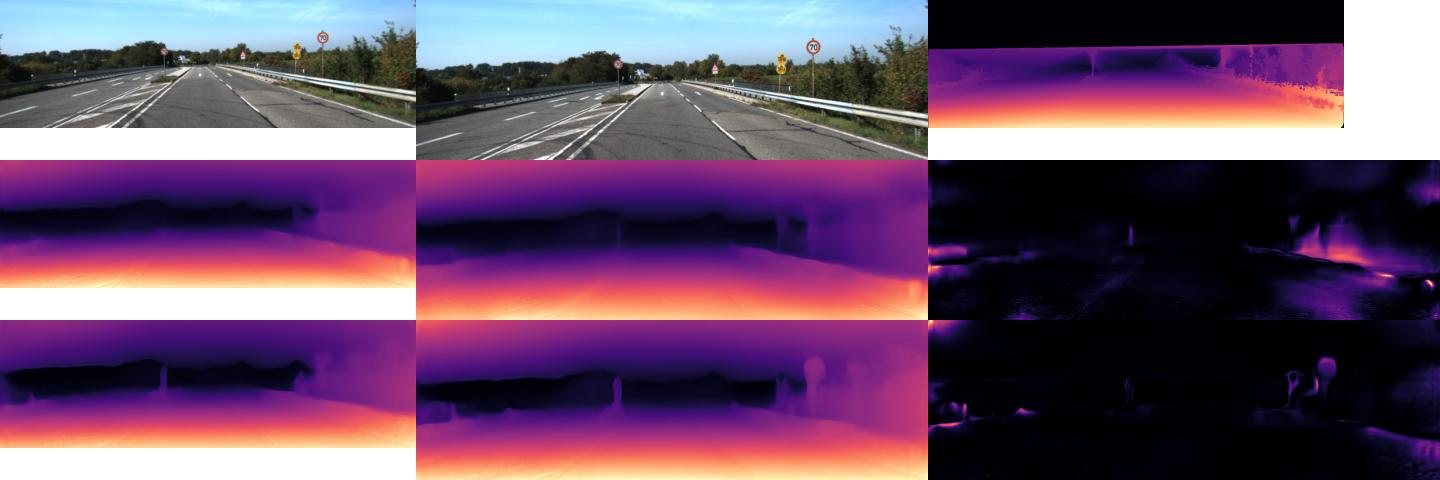}}
	\vspace{-1pt}
	\caption{The influence of $L_{py}$ on the prediction results. `R1': source RGB, transformed RGB, GT. `R2': scheme `w/o $L_{py}$'.}\label{Fig:lpy}
	\vspace{-10pt}
\end{figure*}

Based on the combination scheme obtained from Tab. \ref{Tab:ablation_lpy}, saturation training was conducted for each component of the proposed $L_{py}$, as shown in Tab. \ref{Tab:ablation_pyr_loss}. The results show that the depth prediction performance can be improved by each component of $L_{py}$. Furthermore, using the L1-norm to measure the consistency between feature/disparity maps obtained from $I_{cr}$ and $I_{color}$ is better than using the probability distribution error to measure their consistency with respect to most metrics. This may be because the L1 norm is tighter than the probability distribution error, namely, the L1 norm requires not only the same probability distribution, but also equal values. This constraint makes it easier to enforce consistency between results obtained from $I_{cr}$ and $I_{color}$. Since an interpolation operation is required before the consistency error obtained from samples $I_{cr}$ and $I_{lr}$/$I_{hr}$ is measured, we only force the consistency of the probability distribution between these values.

It can be seen from Fig. \ref{Fig:lpy_illumination} that compared with the scheme `w/o $L_{py}$' (namely, only `$L_{cr}$'), the scheme `$L_{cr}$+$L_{color}$' can improve the robustness of the model to different illuminations. The consistency constraints based on the source image and the flipped image allow the model to obtain left-right consistent predictions and help to obtain sharper boundary predictions, as shown in Fig. \ref{Fig:lpy_flip}. Fig. \ref{Fig:lpy_low} shows that the depth cannot be accurately inferred from some regions of the current resolution (e.g., $128 \times 416$) image by using the weight obtained by training on the current resolution image in some cases, while the corresponding depth can be inferred more accurately from the low resolution (e.g., $96 \times 320$) input image by using the same weight. We believe that this may be because the model can perceive longer-range contextual information from low-resolution input images under the same conditions.   More accurate and consistent depth predictions can be obtained by using the signals calculated from $I_{cr}$ and $I_{lr}$ to guide the model weight update. Compared with the signal calculated from $I_{cr}$,  the signals calculated from $I_{cr}$ and $I_{hr}$ allow the model to focus on more fine-grained information and infer more accurate details, as shown in Fig. \ref{Fig:lpy_high}. Fig. \ref{Fig:hqdec_hqdecv2} shows that the proposed HQDecv2 effectively alleviates the grid artifact phenomenon in HQDec \cite{wangfei_hqdec}, and further improves the quality of the depth map.

In Tab. \ref{Tab:ablation_func_approx_loss}, we conducted ablation studies on the proposed $L_{apx}$ in Sec. \ref{sec:func_approx_loss_method}. Compared with  using only $L_{cr}$ as the supervisory signal, using $L_{apx}$ as the supervisory signal can achieve better depth prediction. This should be because by forcing the output disparity, obtained from SmallDepth, to have the same distribution as the output disparity, obtained from the large model (namely, HQDecv2), it is possible to transfer knowledge from HQDecv2 to SmallDepth.  SmallDepth can learn more accurate knowledge from HQDecv2 if inaccurate prediction areas are masked (e.g., the region where the black hole appears, the region that is farther away). The results also show that the performance of small models can be further improved if the outputs of the encoder and decoder in each stage are also forced to be consistent. More accurate depth prediction results can be obtained if $L_{cr}$ and $L_{apx}$ are used together as supervisory signals.

\section{Conclusions}\label{sec:conclusions}

In this paper, we have presented a fast and accurate self-supervised monocular depth estimation method that (1) utilizes the designed SmallDepth to achieve fast inference; (2) utilizes the proposed ETM to increase the feature representation ability of SmallDepth without changing the complexity of SmallDepth during inference; (3) utilizes the proposed pyramid loss to enhance the perception of different contextual information and robustness to left-right directions and illumination; and (4) utilizes the proposed APX to transfer the knowledge from HQDecv2 into SmallDepth to further improve the accuracy of SmallDepth. The experimental results indicate that the proposed method achieves state-of-the-art results at an inference speed of more than 500 frames per second and with approximately 2 M parameters. However, compared with $L_{cr}$, the $L_{py}$ scheme requires more time and consumes more resources during training because multiple forward and backpropagation steps are needed. Second, the random drop strategy and existing memory saving strategy cannot be utilized at the same time during training. In addition, when there is a large area of sky in a scene, it is often impossible to accurately predict the depth of the area to infinity. We plan to address these problems in future work.

{
	\bibliographystyle{IEEEtran}
	\bibliography{reference}
	
}

\vspace{-2cm}
\begin{IEEEbiography}
	[{\includegraphics[width=0.72in,height=0.9in,clip]{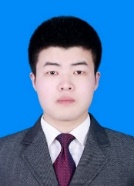}}]
 {Fei Wang} is currently pursuing the Ph.D. degree in University of Chinese Academy of Sciences, Shenzhen Institute of Advanced Technology. His current research interests include computer vision, structure from motion, robotics and deep learning.
\end{IEEEbiography}
\vspace{-2cm}
\begin{IEEEbiography}
	[{\includegraphics[width=0.72in,height=0.9in,clip]{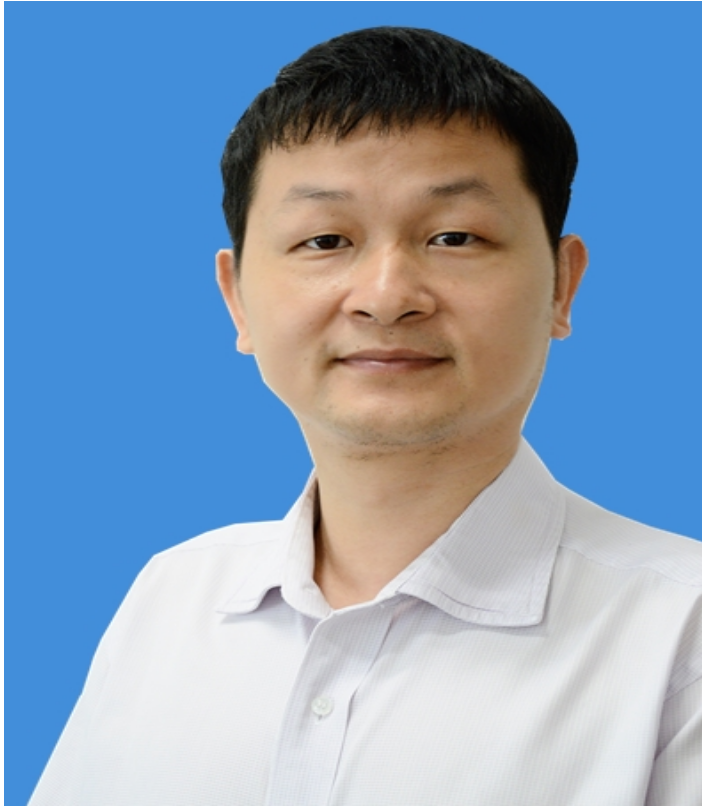}}]
	{Jun Cheng} received the B.Eng. and M.Eng. degrees from the University of Science and Technology of China, Hefei, China, in 1999 and 2002, respectively, and the Ph.D. degree from The Chinese University of Hong Kong, Hong Kong, in 2006. He is currently with the Shenzhen Institute of Advanced Technology, Chinese Academy of Sciences, Shenzhen, China, as a Professor and the Director of the Laboratory for Human Machine Control. His current research interests include computer vision, robotics, machine intelligence, and control.
\end{IEEEbiography}

\end{document}